\RequirePackage[svgnames]{xcolor}
\documentclass[11pt,letterpaper]{ucsdreport}

\usepackage[authoryear,round]{natbib}
\setcitestyle{authoryear,round,citesep={;},aysep={,},yysep={;}}

\usepackage{amsmath,amsfonts,bm}

\def\eqref#1{equation~\ref{#1}}

\def\1{\bm{1}}

\DeclareMathAlphabet{\mathsfit}{\encodingdefault}{\sfdefault}{m}{sl}
\SetMathAlphabet{\mathsfit}{bold}{\encodingdefault}{\sfdefault}{bx}{n}

\usepackage{multirow}
\usepackage{adjustbox}
\usepackage{fontawesome5}
\usepackage{xurl}
\newcommand{\wjepaWebsiteURL}{https://yogesh914.github.io/wjepa}
\newcommand{\wjepaGitHubURL}{https://github.com/Yogesh914/wjepa}
\newcommand{\wjepaHuggingFaceURL}{https://huggingface.co/zou-research-lab/wjepa}
\newcommand{\wjepaResource}[3]{%
  \par\noindent
  \makebox[1.2em][c]{\color{UCSDBlue}#1}\hspace{0.4em}%
  \textbf{#2}:\enspace
  \href{#3}{\expandafter\nolinkurl\expandafter{#3}}\par}

\title{\textbf{W-JEPA}: Wavelet Joint-Embedding Predictive Architecture for Wearable Activity Recognition}
\runningtitle{W-JEPA: Wavelet Joint-Embedding Predictive Architecture for Wearable Activity Recognition}
\hypersetup{
  pdftitle={W-JEPA: Wavelet Joint-Embedding Predictive Architecture for Wearable Activity Recognition},
  pdfauthor={Yogesh Prabhu, Loki Natarajan, Jingjing Zou},
  pdfsubject={Wavelet joint-embedding predictive architecture for wearable activity recognition},
  pdfkeywords={W-JEPA, wavelets, self-supervised learning, wearable activity recognition}
}

\DeclareRobustCommand{\affref}[1]{\hyperlink{affiliation:#1}{#1}}
\DeclareRobustCommand{\correspondref}{\hyperlink{correspondence}{*}}
\author[ ]{%
  Yogesh Prabhu\textsuperscript{\affref{1}},\quad
  Loki Natarajan\textsuperscript{\affref{2}},\quad
  Jingjing Zou\textsuperscript{\affref{2},\affref{3},\affref{4},\correspondref}%
}
\affil[ ]{University of California San Diego}

\authornotes{%
  \footnotetext[1]{\hypertarget{affiliation:1}{}Department of Computer Science and Engineering, University of California San Diego (UCSD).}%
  \footnotetext[2]{\hypertarget{affiliation:2}{}Herbert Wertheim School of Public Health and Human Longevity Science, University of California San Diego (UCSD).}%
  \footnotetext[3]{\hypertarget{affiliation:3}{}UCSD Shiley-Marcos Alzheimer's Disease Research Center.}%
  \footnotetext[4]{\hypertarget{affiliation:4}{}UCSD Moores Cancer Center.}%
  \begingroup
  \footnotetext[0]{\hypertarget{correspondence}{}\textit{Corresponding author(s):} Jingjing Zou (\href{mailto:j2zou@ucsd.edu}{j2zou@ucsd.edu}).}%
  \endgroup
  \setcounter{footnote}{4}%
}

\begin{document}
\begin{abstract}

A major challenge in wearable device activity recognition is the heterogeneity across datasets in sensor placement, channel count, sampling rate and subjects. Moreover, the motion patterns that distinguish activities often appear as subtle, scale-specific variations within otherwise similar waveforms. As a result, raw signals are often ambiguous, and models trained on them tend to rely on cues specific to a particular sensor setup or participant, which limits both cross-dataset transfer and generalization to unseen participants. Addressing this requires representations that preserve such activity-relevant, scale-specific variation. We introduce W-JEPA, a self-supervised framework that exploits the multiscale structure of inertial motion and leverages a joint-embedding predictive architecture, by organizing windows into a fixed grid of wavelet tokens indexed by scale, sensor axis, and time. These wavelet tokens help expose activity-specific detail within broadly similar raw signals, and serve as the units of masked latent prediction. W-JEPA predicts masked token representations supplied by an exponential-moving-average (EMA) target encoder. Because recovering a masked token requires relating motion at one scale and axis to motion at other scales and axes, the model learns the cross-scale, cross-axis structure of human movement. Across six wearable activity benchmarks, W-JEPA outperforms state-of-the-art methods under random-window splits and remains the strongest when subjects are held out, improving average macro-F1 by 8.7 percentage
points over the strongest published baseline. These results indicate that physically grounded multiscale tokenization, combined together with latent predictive learning, provides a promising foundation for transferable representations of heterogeneous wearable inertial data.

\vspace{3mm}
\wjepaResource{\faGlobe}{Website}{\wjepaWebsiteURL}
\wjepaResource{\faGithub}{Code Repository}{\wjepaGitHubURL}
\wjepaResource{\raisebox{-0.18em}{\includegraphics[height=1.15em]{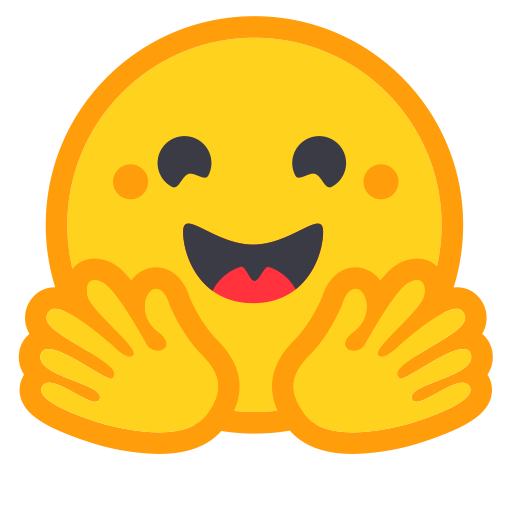}}}{Hugging Face}{\wjepaHuggingFaceURL}
\end{abstract}

\maketitle

\section{Introduction}
\label{sec:intro}

Public wearable datasets provide substantial unlabeled recordings for learning representations of human motion, but they differ in sensor placement, channel count, sampling characteristics, participants, and activity vocabularies \citep{chen2021deep}. A useful representation must therefore capture which activity is being performed, independent of how and on whom it was recorded. This is difficult because the information that separates activities is often small. Similar physical activities can produce nearly the same rhythmic, periodic pattern in raw sensor recordings and differ mainly in the finer temporal structure within each movement cycle \citep{s21196393, tian2020wavelet}. Because this detail is small relative to differences in sensor setup and individual movement style, a model trained on raw signals can fit its training data using cues tied to the setup or to particular participants. Such cues support strong performance on familiar participants but may not carry over to people unseen during training. In our experiments, every method loses at least 14 points of average macro-F1 when evaluation moves from random windows to held-out participants (\S\ref{sec:main_results}). Broad movement patterns and fine activity-specific detail, however, tend to occupy different timescales. A representation that separates motion by timescale can keep this detail from being masked by larger variation in the signal, while still allowing broad patterns to inform recognition.

\begin{figure}[t]
    \centering
    \begin{minipage}[t]{0.28\textwidth}
        \centering
        \textbf{(a)}\par\smallskip
        \includegraphics[width=\linewidth]{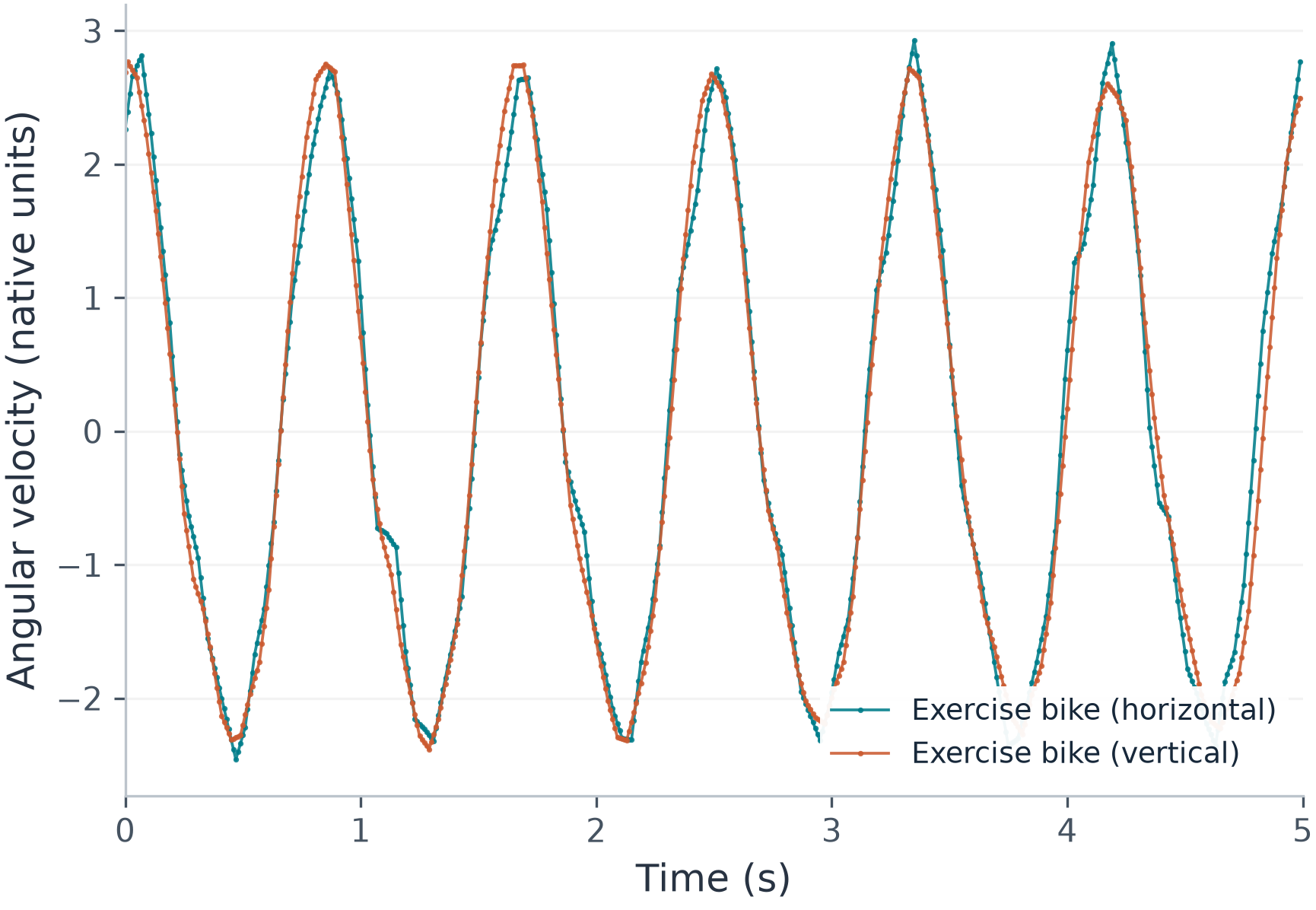}
    \end{minipage}\hspace{0.02\textwidth}%
    \begin{minipage}[t]{0.286\textwidth}
        \centering
        \textbf{(b)}\par\smallskip
        \includegraphics[width=\linewidth]{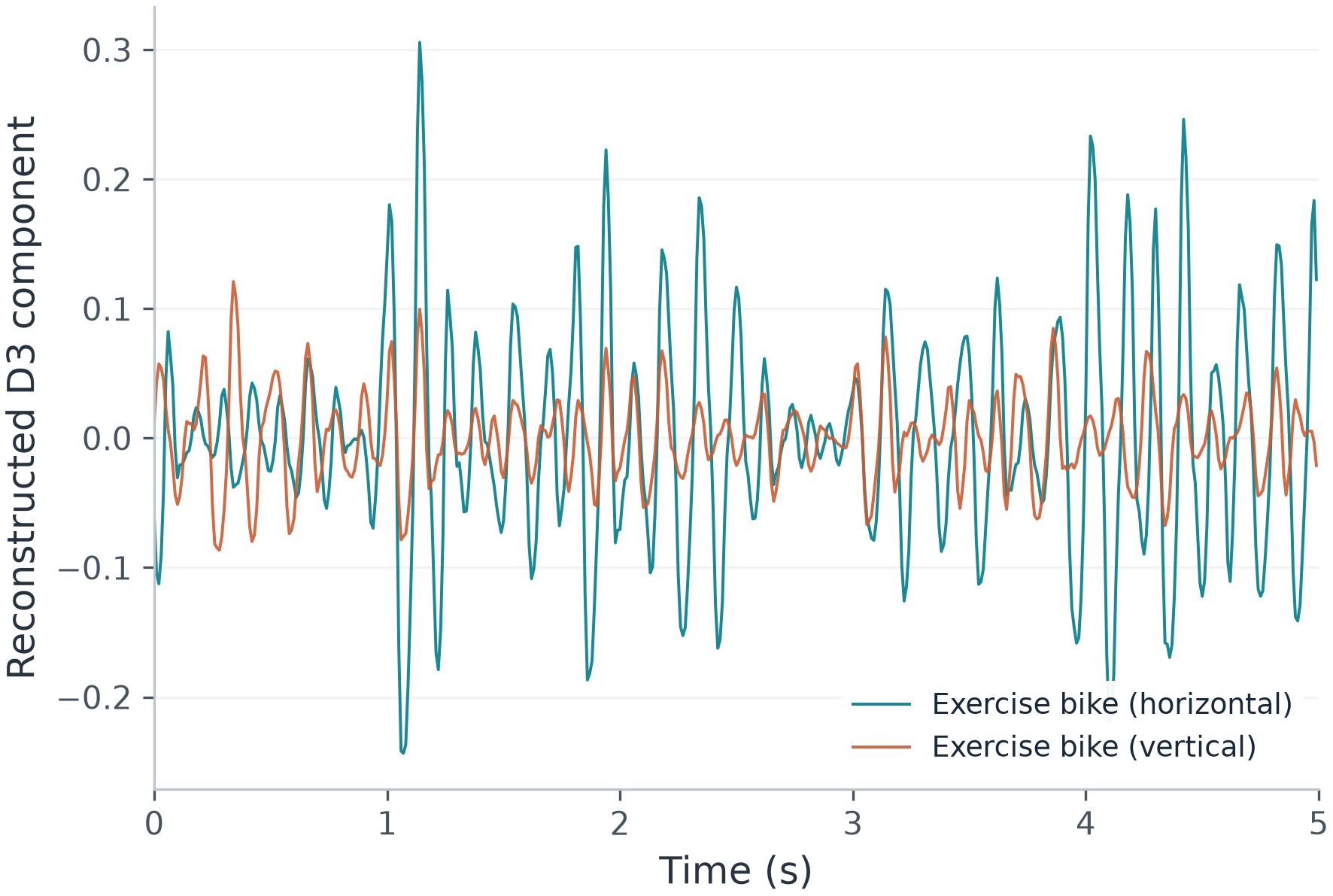}
    \end{minipage}\hspace{0.02\textwidth}%
    \begin{minipage}[t]{0.384\textwidth}
        \centering
        \textbf{(c)}\par\smallskip
        \includegraphics[width=\linewidth]{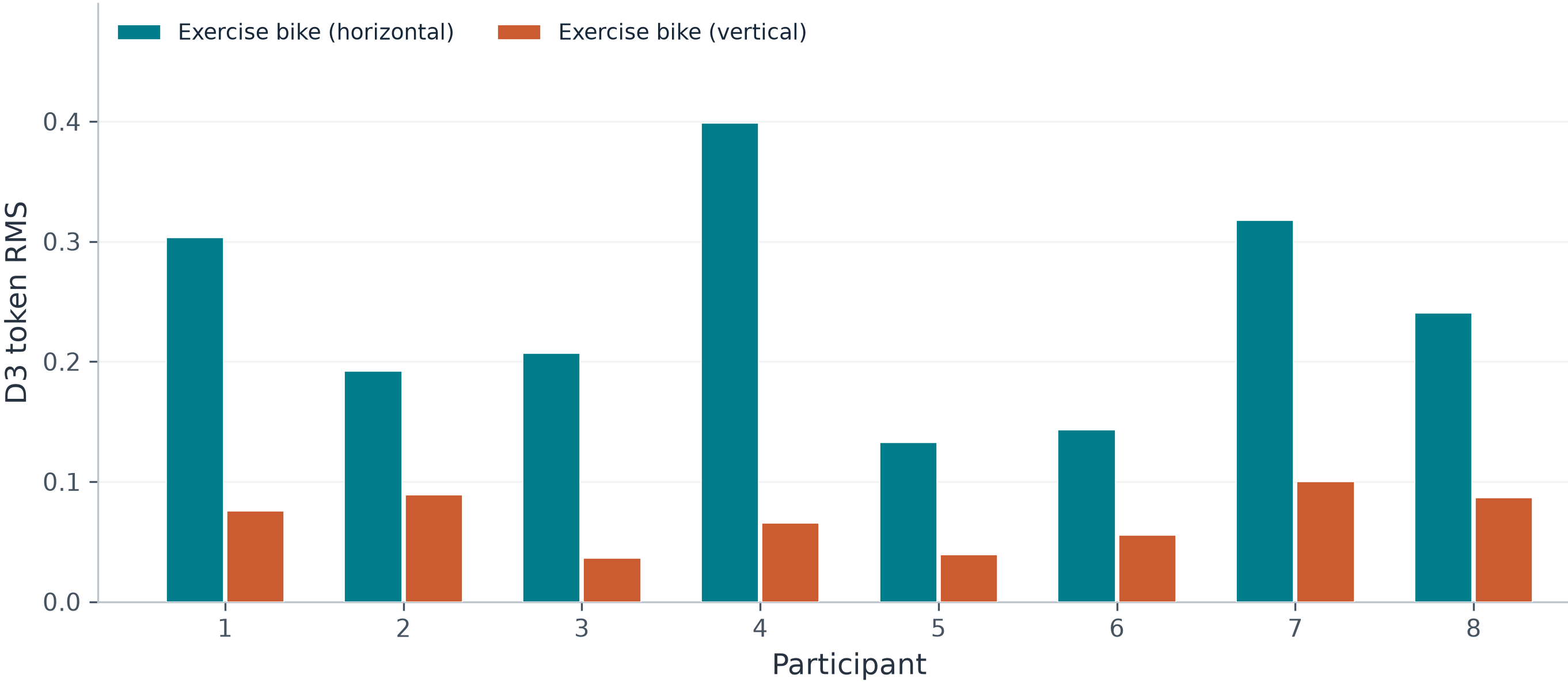}
    \end{minipage}
    \caption{\textbf{Wavelet tokens reveal differences between similar activity signals.}
    \textbf{(a)} Sample window of horizontal and vertical exercise-bike cycling from a DSADS participant's left leg gyroscope $z$-axis exhibit similar raw waveforms.
    \textbf{(b)} The reconstructed db4 $D_3$ component reveals larger pronounced oscillations during horizontal cycling in this pair.
    \textbf{(c)} Mean RMS of the second $D_3$ token across 59 windows per activity and participant, with higher values for horizontal cycling in all eight participants. More details appear in Appendix~\ref{sec:dsads_detail}.}
    \label{fig:activity_detail}
\end{figure}

Figure~\ref{fig:activity_detail} shows an example of two labels from the DSADS dataset that are easy to confuse. Panel (a) shows horizontal and vertical exercise-bike cycling from the same participant. The two raw waveforms nearly overlap, with a Pearson correlation of 0.992. Their reconstructed $D_3$ wavelet components, however, reveal pronounced oscillations during horizontal cycling (panel b). Across all eight participants, the $D_3$ token has higher mean RMS for horizontal cycling than for vertical cycling (panel c). The information that separates the two activities is present in the signal and becomes visible once this scale is isolated. Self-supervised pretraining could learn such distinctions from unlabeled recordings, but masked-modeling approaches for wearables typically hide and predict raw temporal patches \citep{ryu2025moca, noor2026joint}, which mix all timescales within each interval. This raises the question we study: \emph{how should the units of latent prediction be defined to preserve activity-specific detail while learning transferable motion representations from heterogeneous wearable signals?}

We propose W-JEPA, which answers this question by using fixed wavelet tokens as the units of masked latent prediction. A discrete wavelet transform separates each sensor axis into localized components at individual timescales \citep{daubechies1992ten}, and W-JEPA summarizes each resulting band with tokens indexed by scale, sensor axis, and time (Figure~\ref{fig:method}a). The tokenizer is fixed, meaning no part of it is fit to the pretraining datasets. Because all recordings are resampled to a common rate, each band covers the same nominal frequency range in every dataset, so a token at a given scale carries the same physical meaning across sensor setups. The token grid grows with the number of channels, and only a learned channel embedding depends on the sensor layout. Datasets with different layouts therefore share a common set of prediction units and differ only in how many channels they contribute.

During pretraining, a random subset of tokens is masked. A context encoder processes the visible tokens, and a predictor estimates the latent features that an EMA target encoder assigns to the masked positions (Figure~\ref{fig:method}b). Predicting in latent space means the model is not asked to reproduce raw sample values, whose amplitude and phase can vary with participant and sensor placement. Masks are drawn independently across channels and bands, so recovering a masked token encourages the model to relate motion at one scale on one axis to motion at other scales on other axes. Because each scale has its own tokens, fine activity-specific detail becomes a prediction target in its own right instead of a small component of a patch dominated by larger variation.

We evaluate W-JEPA on six wearable activity benchmarks. For each target dataset, we pretrain on the remaining five and fine-tune on two to six of its participants. W-JEPA achieves the highest mean accuracy and macro-F1 on every dataset under both subject-disjoint and random-window splits. With participants held out, its average macro-F1 is 80.6, a 12.1\% relative improvement over the strongest published baseline at 71.9. The wavelet tokenizer accounts for most of this margin. Replacing it with raw temporal patches lowers average macro-F1 by 7.6 points under participant holdout, compared with 2.9 points under random windows. The tokenizer therefore contributes more than twice as much when test participants are unseen, consistent with wavelet tokens reducing reliance on participant-specific cues.

Our contributions are as follows:
\begin{itemize}
    \item \textbf{Wavelet tokens as prediction units.} A fixed tokenizer that maps each inertial window to tokens indexed by scale, sensor axis, and time. Each band covers the same frequency range in every dataset, and only a channel embedding depends on the sensor layout. Prior work uses wavelets as classification features, forecasting tokens,
or reconstruction targets, W-JEPA uses wavelet tokens as units of masked latent
prediction over multi-axis inertial data (\S\ref{sec:tokenizer}).

    \item \textbf{W-JEPA.} A masked latent-prediction framework over these tokens. Masks are drawn independently across channels and bands, so predicting a masked token requires relating motion across scales and axes without reconstructing raw signal values (\S\ref{sec:jepa}).
    \item \textbf{Cross-dataset evaluation under participant holdout.} With each target excluded from pretraining, W-JEPA outperforms eight baselines on six benchmarks under both subject-disjoint and random-window splits, and it loses the least performance when test participants are unseen (\S\ref{sec:main_results}). Its lead holds when fine-tuning is restricted to wrist sensors (\S\ref{sec:restricted_settings}).
    \item \textbf{Evidence that the wavelets matter.} Replacing wavelet tokens with raw temporal patches lowers subject-disjoint macro-F1 on every dataset, especially when few training participants are available (\S\ref{sec:tokenizer_analysis}). Without wavelet tokens, 41.5\% of horizontal exercise-bike windows are classified as vertical cycling, compared with 0.8\% with them (\S\ref{sec:dsads_cycling}).
\end{itemize}

\section{Related Work}
\label{sec:related}

Self-supervised pretraining for wearables has explored many training objectives, but the unit of prediction has received less attention. Existing units are either contiguous intervals of raw samples, which mix all motion scales, or codes learned from the pretraining data, whose meaning depends on that data. W-JEPA uses units that are fixed in advance and separated by scale. We review the work this choice builds on.

\paragraph{Wavelets for activity recognition.}
Wavelets separate nonstationary signals into components localized in both time
and scale \citep{daubechies1992ten}, whereas Fourier components, used for
example in the adaptive spectral blocks of TSLANet \citep{eldele2024tslanet},
extend over the whole window. Wavelet energy features and wavelet inputs to convolutional networks have long been used for supervised activity recognition \citep{tian2020wavelet,vuong2023wavelet}. In
forecasting, WaveToken quantizes wavelet coefficients into a vocabulary for
autoregressive prediction \citep{masserano2024enhancing}. In self-supervised
pretraining, \citet{thukral2026wavelet} reconstructs masked wavelet coefficients
of PPG signals, which still requires reproducing coefficient values whose
amplitudes vary across participants. These methods use wavelets
as features, forecasting tokens, or reconstruction targets. To our knowledge,
prior work has not used wavelets to define the units of masked latent
prediction for multi-axis inertial data. Each token is addressed by scale, sensor axis, and time, so a mask
can hide one scale on one axis while others stay visible, and the model
predicts learned representations rather than coefficients.

\paragraph{Prediction units and targets in wearable self-supervision.}
Contrastive and bootstrap methods relate augmented views of a window \citep{eldele2021time,grill2020bootstrap,guo2024modality,zhang2022tfc}. Their performance depends strongly on the choice of augmentation \citep{qian2022makes}, which can remove the fine detail that separates similar activities. Masked modeling makes the unit explicit. Most wearable methods reconstruct masked sensor values or temporal patches \citep{zerveas2021transformer,haresamudram2020masked,xu2021limubert,ryu2025moca}, which requires reproducing amplitude and phase details that vary with participant and sensor placement. MoPFormer instead predicts codes from a learned vocabulary of motion primitives \citep{zhang2026mopformer}, so its units depend on the pretraining data. Joint-embedding predictive architectures replace reconstruction with the prediction of learned representations \citep{lecun2022path,assran2023self,bardes2024revisiting} and have been extended to time series, skeletal actions, and EEG \citep{ennadir2025joint,abdelfattah2024s,guetschel2024s}. HAR-JEPA brings this idea to activity recognition with temporal-patch units \citep{noor2026joint}. Latent targets remove the burden of signal reconstruction, but temporal patches still mix all motion scales within each interval. W-JEPA keeps the latent target and replaces the unit with wavelet tokens indexed by scale, sensor axis, and time.

\paragraph{Generalization across datasets and participants.}
Differences in sensor placement, device, sampling protocol, and participant behavior limit transfer in wearable activity recognition \citep{chen2021deep,lu2023diversify}. Recent methods address this through physics-informed augmentation \citep{xu2023unihar}, hierarchical pretraining with augmentation \citep{hong2024crosshar}, or alignment with language \citep{zhang2024unimts,ding2026halo}. These methods change the training signal. W-JEPA changes the unit of representation, so that a token at a given scale covers the same frequency band in every dataset, and the two directions are complementary. Evaluation matters as well. Placing windows from the same participant in both training and test sets overstates performance \citep{hammerla2015lets}. We therefore treat participant-disjoint results as the primary evaluation, and we also exclude each target dataset from pretraining (\S\ref{sec:main_results}).

\section{Method}
\label{sec:methods}
\begin{figure}[ht]
    \centering
    \includegraphics[width=0.90\textwidth]{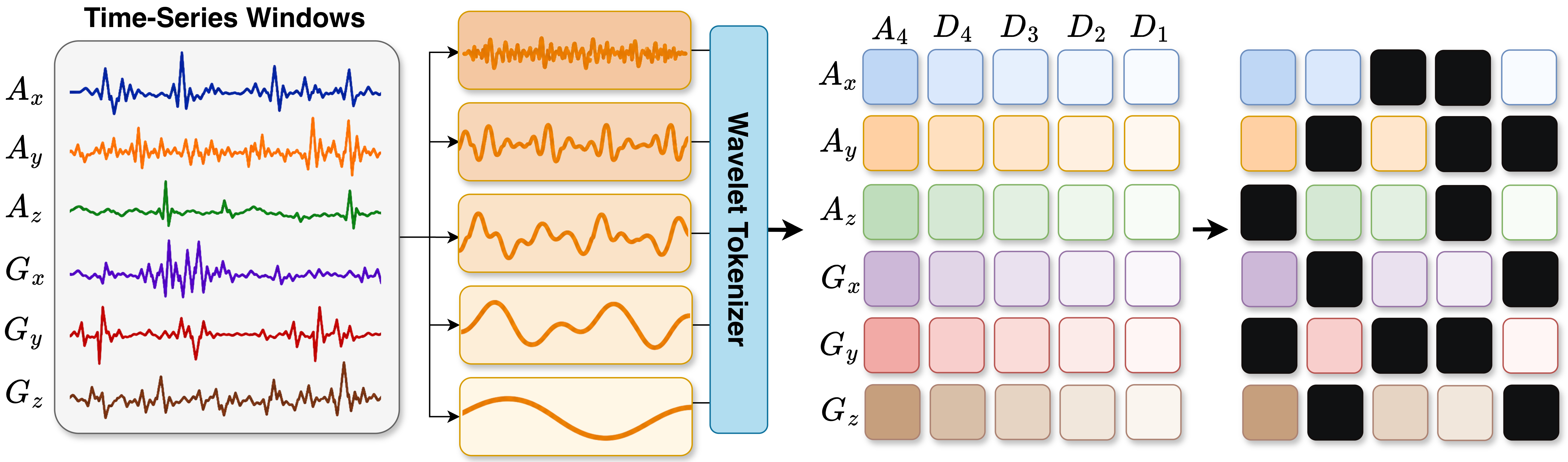}\\[1pt]
    {\small (a) Wavelet tokenization and masking}\\[5pt]
    \includegraphics[width=0.90\textwidth]{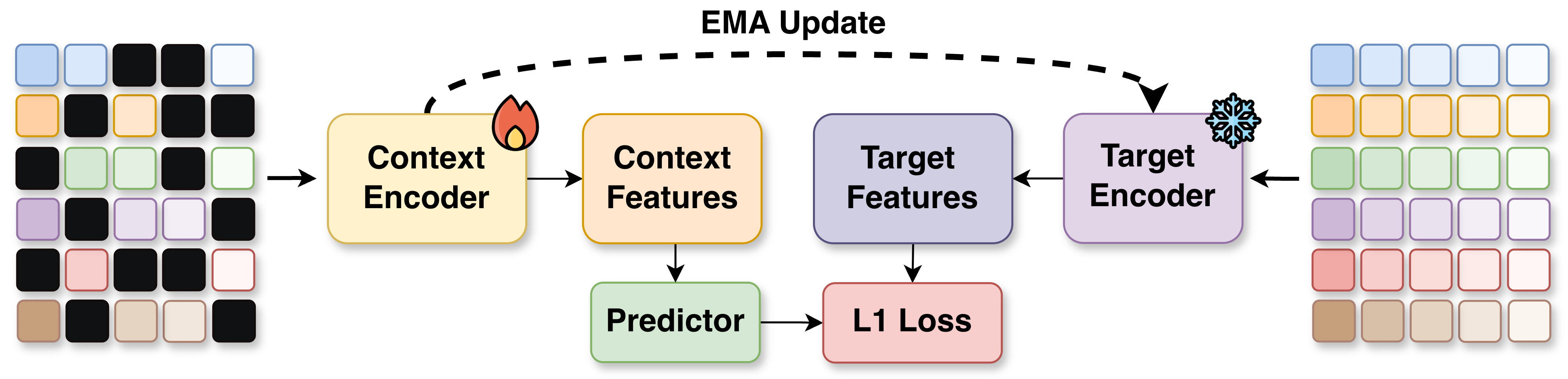}\\[1pt]
    {\small (b) Masked latent prediction}
    \caption{\textbf{W-JEPA.} (a) Wavelet tokenization and masking. Rows are sensor channels and columns are wavelet bands $A_4$--$D_1$, ordered from coarse to fine. Each channel and wavelet band combination produces two temporal tokens, shown as a single cell for clarity and masked together; black cells indicate masked token pairs.
    (b) Masked latent prediction. The predictor estimates representations
at masked positions from visible-token context, with targets supplied
by an EMA target encoder.}
    \label{fig:method}
\end{figure}

Figure~\ref{fig:method} summarizes W-JEPA. Given an inertial window
$x\in\mathbb{R}^{C\times T}$ with $C$ sensor channels (one per sensor axis) and
$T$ samples, a fixed wavelet tokenizer maps $x$ to a grid of tokens indexed by
channel, wavelet band, and time bin (\S\ref{sec:tokenizer}). During pretraining,
a random subset of channel--band token pairs are masked. A context encoder reads
the visible tokens, a predictor estimates a latent target at each masked
position, and an EMA target encoder computes those targets from the full token
sequence. The encoders, predictor, and EMA update follow standard JEPA
\citep{lecun2022path,assran2023self,bardes2024revisiting}. 

The tokenizer supplies the structured prediction space. As mentioned before,
Figure~\ref{fig:activity_detail} shows that detail bands expose fluctuations
between hard to distinguish labels. Summarizing each wavelet band separately makes these finer details available to the encoder alongside lower-frequency signal components. Masked prediction can then use relationships among the visible
scales, channels, and time bins. This is the motivation for the design; the
tokenizer comparison in \S\ref{sec:experiments} evaluates its effect on
recognition performance. The design choices follow from this view. W-JEPA's changes lie
in the tokenizer (\S\ref{sec:tokenizer}) and in the masking scheme and objective
(\S\ref{sec:jepa}).

\subsection{Wavelet tokenization}
\label{sec:tokenizer}

An inertial window can contain slow variations, rhythmic motion, and brief
transients. A discrete wavelet transform separates these timescales while retaining temporal localization
\citep{daubechies1992ten}. Tokenizing along those bands means every token
arrives with a physically meaningful address: a scale, a sensor axis, and a
moment in time.

\paragraph{Channel and band axes.}
Let $\mathcal{C}=\{1,\ldots,C\}$ index the available sensor channels.
Figure~\ref{fig:method}a illustrates the common six-channel case
$\{A_x,A_y,A_z,G_x,G_y,G_z\}$, where $A$ and $G$ denote accelerometer and
gyroscope channels. For every $c\in\mathcal{C}$, a four-level Daubechies-4
discrete wavelet transform with reflective boundary extension produces four
detail-coefficient sequences $d_{c,1},\ldots,d_{c,4}$ and one
approximation-coefficient sequence $a_{c,4}$ \citep{wolter2024ptwt}. We call
these five sequences \emph{wavelet bands} and index them from fine to coarse. The wavelet bands provide a fixed organization of the signal by
timescale. Within each band, RMS summarizes coefficient magnitude,
while the mean retains signed coefficient information. These
statistics provide scale-specific descriptors for token construction.

\paragraph{Token construction.}
Each of the five coefficient sequences is divided into $P$ contiguous bins of
approximately equal length, so that every band yields the same number of bins
regardless of how many coefficients it holds. Let $q_{ckb}$ hold the
coefficients in bin $b\in\{1,\ldots,P\}$ of band $k\in\{1,\ldots,5\}$ for
channel $c$. We summarize each bin by two statistics, its mean $\mu_{ckb}$ and
its RMS magnitude $r_{ckb}$:
\begin{equation}
\begin{aligned}
    \mu_{ckb} &= \frac{1}{|q_{ckb}|}\sum_{v\in q_{ckb}}v, \qquad
    r_{ckb} = \sqrt{\frac{1}{|q_{ckb}|}\sum_{v\in q_{ckb}}v^2 + 10^{-6}}.
\end{aligned}
    \label{eq:descriptor}
\end{equation}
Together, these summarize scale-specific signal magnitude and preserve signed coefficient information. We use $P=2$ bins per band.
A two-layer MLP $g:\mathbb{R}^{2}\rightarrow\mathbb{R}^{384}$ maps this pair to
a content embedding, and learned embeddings of the channel, band, and bin
indices supply the token's address:
$u_{ckb}=g([\mu_{ckb},r_{ckb}])+e^{\mathrm{ch}}_c
+e^{\mathrm{band}}_k+e^{\mathrm{bin}}_b$.
Flattening the grid gives a sequence of $N=5CP$ tokens, in which each position
corresponds to exactly one triple $(c,k,b)$. Of the learned components in the token embedding, only $e^{\mathrm{ch}}_c$ depends on the sensor layout,
which is what makes the adaptation in downstream finetuning cheap.

\subsection{Masked latent pretraining}
\label{sec:jepa}

W-JEPA masks after tokenization and predicts learned representations of the
masked tokens. The latent targets allow the training objective to operate on
features learned by the target encoder, with no raw-sample reconstruction
term. For window $i$, let $\mathcal{M}_i$ denote the masked
token positions and $\mathcal{V}_i$ their visible complement.

\paragraph{Asynchronous paired-token masking.}
Each channel and wavelet-band combination produces two temporal tokens.
We group these tokens into $G=N/2$ disjoint pairs
$\{\mathcal{P}_g\}_{g=1}^{G}$, where each pair contains the two tokens
from the same channel and wavelet band. Given a mask ratio $\rho$,
we uniformly sample a subset $\mathcal{S}_i$ of
$|\mathcal{S}_i|=\lfloor\rho G\rfloor$ pairs, with a fresh draw per
window, and mask both tokens in each selected pair, yielding
$\mathcal{M}_i=\bigcup_{g\in\mathcal{S}_i}\mathcal{P}_g$.
Masking remains asynchronous across channels and bands: selecting one
pair does not require masking other channels or bands.
This preserves complementary information in the visible context and
encourages the predictor to exploit dependencies across sensors, axes, and wavelet scales. MoCA similarly finds that independently
masking sensor axes improves reconstruction pretraining relative
to synchronized masking \citep{ryu2025moca}. Within each
pair, masking both temporal tokens prevents the predictor from relying
on the other token from that same pair.
Figure~\ref{fig:method}a illustrates our paired-token mask.

\paragraph{Objective.}
Let $p_{ij}\in\mathbb{R}^{D}$ be the predictor output and $\bar{z}_{ij}\in\mathbb{R}^{D}$ the normalized latent target at masked token position $j$. For batch size $B$, the prediction loss is
\begin{equation}
    \mathcal{L}_{\mathrm{pred}}
    =
    \frac{1}{B}
    \sum_{i=1}^{B}
    \frac{1}{|\mathcal{M}_i|D}
    \sum_{j\in\mathcal{M}_i}
    \left\|p_{ij}-\bar{z}_{ij}\right\|_1.
    \label{eq:prediction}
\end{equation}

To discourage representation collapse, we add the variance hinge from VICReg without its covariance term \citep{bardes2022vicreg}. Let $\sigma_{ir}$ be the sample standard deviation across predictions $\{p_{ij}:j\in\mathcal{M}_i\}$ in feature coordinate $r$. We define
\begin{equation}
    \mathcal{L}_{\mathrm{var}}
    =
    \frac{1}{BD}
    \sum_{i=1}^{B}
    \sum_{r=1}^{D}
    \max(0,1-\sigma_{ir}),
    \qquad
    \mathcal{L}_{\mathrm{pre}}
    =
    \mathcal{L}_{\mathrm{pred}}
    + 0.0625\,\mathcal{L}_{\mathrm{var}}.
    \label{eq:objective}
\end{equation}

The spread is computed within each window over its own masked positions, so the
hinge asks that a single window's predictions vary across scales and axes.
In a diagnostic comparison, substituting
SIGReg \citep{balestriero2025lejepa} while holding the rest of the pipeline
fixed costs accuracy (Appendix Table~\ref{tab:ablations}). For downstream fine-tuning, a target dataset with $C_{\mathrm{tar}}$ channels, the same tokenizer
produces $N=5C_{\mathrm{tar}}P$ token positions. The target encoder and
predictor are discarded, and the context encoder is fine-tuned end-to-end with
a classification head on all available channels.

\section{Experiments}
\label{sec:experiments}

We ask three questions. Does W-JEPA, pretrained on other datasets, recognize
activities of participants unseen during fine-tuning?
Does its advantage hold when the target sensor layout is restricted
(\S\ref{sec:restricted_settings})? And what do wavelet tokens contribute beyond
the JEPA objective (\S\ref{sec:tokenizer_analysis})?

\subsection{Main Results}
\label{sec:main_results}

\paragraph{Datasets and pretraining pool.}
We evaluate on PAMAP2~\citep{pamap2_physical_activity_monitoring_231,reiss2012pamap2}, DSADS~\citep{daily_and_sports_activities_256,altun2010dsads}, MHealth~\citep{banos2014mhealth,banos2015design}, RealWorld~\citep{sztyler2016realworld}, UCI-HAR~\citep{anguita2013ucihar,Anguita2013APD}, and USC-HAD~\citep{zhang2012uschad}. Appendix~\ref{sec:dataset_details} describes each dataset. For each target, the remaining five datasets form the unlabeled pretraining pool, matching MoPFormer's setup \citep{zhang2026mopformer}. Each source contributes the same six-channel view: the $x$, $y$, and $z$ axes from one accelerometer and one gyroscope, while downstream fine-tuning uses all configured target channels. All recordings are resampled to 100 Hz and segmented into non-overlapping windows of $T=500$ samples (5 seconds). We do not apply channel normalization.

\paragraph{Evaluation splits.}
Our primary evaluation uses subject-disjoint splits: all windows from a
participant belong exclusively to either downstream fine-tuning or testing.
The target dataset is also excluded from pretraining. For comparison, we
evaluate all methods under a random-window protocol that assigns 20\% of
windows within each activity class to training and 80\% to evaluation; this
protocol can place the same participant in both partitions. We report accuracy
and macro-F1 over five random seeds, with dataset averages weighted equally.
Sample and participant counts for the subject-disjoint splits are given in
Appendix Table~\ref{tab:dataset_statistics}.

\paragraph{Baselines.}
We compare with eight baselines, all re-run with the same
preprocessing, splits, and seeds: BYOL \citep{grill2020bootstrap} with the
permutation-and-jitter (perm\_jit) and low-frequency-component (lfc)
augmentations from \citet{qian2022makes} and with temporal mixup
\citep{10177231}; ModCL, a contrastive method guided by modality consistency
\citep{guo2024modality}; MoPFormer, which also pretrains across datasets
\citep{zhang2026mopformer}; CALANet, an efficient HAR architecture
\citep{park2024calanet}; TSLANet, a general time-series backbone
\citep{eldele2024tslanet}; and HAR-JEPA, applies JEPA to wearable activity recognition \citep{noor2026joint}. W-JEPA (no wavelets), described in
\S\ref{sec:tokenizer_analysis}, isolates the tokenizer.

\paragraph{Results.} Table~\ref{tab:subject_results} evaluates recognition on participants excluded
from training and Table~\ref{tab:main_results} reports the random-window evaluation. W-JEPA achieves the
highest mean accuracy and macro-F1 on every dataset under both setups,
although its margin varies by dataset.
Comparing these tables shows how
much random-window evaluation overstates performance on new people. TSLANet falls the most, while HAR-JEPA remains the strongest baseline. W-JEPA falls by the least of any method.

\begin{table}[ht]
\caption{Subject-disjoint comparison across six wearable activity recognition
datasets. Results are averaged over five random seeds under the unified protocol of
\S\ref{sec:experiments}.}
\label{tab:subject_results}
\centering
\setlength{\tabcolsep}{3.5pt}
\renewcommand{\arraystretch}{1.05}
\small
\resizebox{\textwidth}{!}{%
\begin{tabular}{@{}llrrrrrr@{}}
\toprule
Method & Metric & PAMAP2 & DSADS & MHealth & RealWorld & UCI-HAR & USC-HAD \\
\midrule
\multirow{2}{*}{BYOL + perm\_jit}
& Acc. & $65.65 \pm 3.68$ & $64.66 \pm 5.59$ & $57.70 \pm 19.68$ & $67.59 \pm 0.53$ & $74.49 \pm 2.78$ & $57.81 \pm 3.15$ \\
& F1   & $65.33 \pm 4.65$ & $63.99 \pm 6.12$ & $51.32 \pm 20.57$ & $62.99 \pm 0.45$ & $71.43 \pm 2.98$ & $53.71 \pm 2.69$ \\
\midrule

\multirow{2}{*}{BYOL + lfc}
& Acc. & $72.01 \pm 3.32$ & $65.13 \pm 3.27$ & $73.10 \pm 1.25$ & $65.42 \pm 1.37$ & $77.36 \pm 3.48$ & $59.48 \pm 3.70$ \\
& F1   & $72.16 \pm 3.56$ & $63.79 \pm 4.64$ & $73.46 \pm 0.51$ & $59.87 \pm 1.51$ & $76.21 \pm 3.76$ & $57.63 \pm 2.71$ \\
\midrule

\multirow{2}{*}{BYOL + Mixup}
& Acc. & $72.94 \pm 2.00$ & $65.74 \pm 5.73$ & $74.13 \pm 3.41$ & $64.31 \pm 1.74$ & $75.82 \pm 4.68$ & $58.09 \pm 1.45$ \\
& F1   & $73.30 \pm 2.08$ & $64.25 \pm 6.91$ & $73.76 \pm 3.77$ & $58.92 \pm 1.99$ & $72.89 \pm 5.86$ & $56.43 \pm 1.09$ \\
\midrule

\multirow{2}{*}{ModCL}
& Acc. & $72.52 \pm 2.18$ & $61.00 \pm 1.55$ & $69.53 \pm 5.47$ & $66.25 \pm 2.64$ & $81.35 \pm 5.43$ & $55.75 \pm 5.69$ \\
& F1   & $72.32 \pm 2.79$ & $59.83 \pm 2.03$ & $68.16 \pm 6.12$ & $64.92 \pm 2.04$ & $81.02 \pm 5.45$ & $55.06 \pm 5.19$ \\
\midrule

\multirow{2}{*}{TSLANet}
& Acc. & $65.85 \pm 1.08$ & $70.03 \pm 8.61$ & $77.55 \pm 3.00$ & $57.95 \pm 1.88$ & $80.19 \pm 2.32$ & $60.91 \pm 1.53$ \\
& F1   & $66.43 \pm 1.27$ & $68.48 \pm 9.20$ & $76.82 \pm 3.42$ & $54.67 \pm 1.95$ & $80.10 \pm 2.33$ & $57.40 \pm 0.61$ \\
\midrule

\multirow{2}{*}{CALANet}
& Acc. & $69.40 \pm 2.34$ & $63.16 \pm 5.53$ & $86.37 \pm 2.26$ & $64.51 \pm 2.60$ & $81.86 \pm 1.00$ & $63.58 \pm 2.59$ \\
& F1   & $69.11 \pm 2.29$ & $61.78 \pm 5.97$ & $85.24 \pm 1.90$ & $61.04 \pm 2.87$ & $81.38 \pm 0.90$ & $59.43 \pm 1.96$ \\
\midrule

\multirow{2}{*}{MoPFormer}
& Acc. & $72.43 \pm 1.12$ & $61.10 \pm 9.66$ & $81.42 \pm 1.16$ & $67.38 \pm 2.74$ & $80.93 \pm 5.52$ & $60.70 \pm 4.62$ \\
& F1   & $69.40 \pm 1.09$ & $57.90 \pm 10.76$ & $81.23 \pm 1.03$ & $68.60 \pm 3.41$ & $78.16 \pm 6.43$ & $58.76 \pm 4.04$ \\
\midrule

\multirow{2}{*}{HAR-JEPA}
& Acc. & $61.08 \pm 0.89$ & $73.16 \pm 0.98$ & $87.38 \pm 1.89$ & $64.94 \pm 0.71$ & $81.57 \pm 1.31$ & $61.03 \pm 2.75$ \\
& F1   & $63.53 \pm 0.76$ & $73.59 \pm 0.88$ & $87.64 \pm 1.80$ & $64.50 \pm 0.89$ & $81.99 \pm 0.29$ & $60.22 \pm 1.17$ \\
\midrule

\multirow{2}{*}{W-JEPA (no wavelets)}
& Acc. & $81.34 \pm 0.61$ & $70.00 \pm 1.95$ & $73.33 \pm 2.13$ & $71.06 \pm 1.64$ & $82.71 \pm 3.40$ & $59.49 \pm 0.91$ \\
& F1   & $81.64 \pm 0.31$ & $69.98 \pm 2.51$ & $73.68 \pm 2.68$ & $71.48 \pm 1.00$ & $82.00 \pm 3.47$ & $59.52 \pm 0.53$ \\
\midrule

\multirow{2}{*}{\textbf{W-JEPA}}
& Acc. & $\boldsymbol{86.96 \pm 0.62}$ & $\boldsymbol{82.87 \pm 1.49}$ & $\boldsymbol{89.68 \pm 1.09}$ &
$\boldsymbol{80.04 \pm 0.41}$ & $\boldsymbol{84.70 \pm 1.94}$ & $\boldsymbol{64.35 \pm 3.28}$ \\
& F1   & $\boldsymbol{87.19 \pm 0.61}$ & $\boldsymbol{82.76 \pm 1.76}$ & $\boldsymbol{88.89 \pm 1.24}$ &
$\boldsymbol{80.53 \pm 1.10}$ & $\boldsymbol{83.97 \pm 2.08}$ & $\boldsymbol{60.35 \pm 1.40}$ \\
\bottomrule
\end{tabular}%
}

\end{table}

\begin{table}[ht]
\caption{Random-window comparison across six wearable activity recognition
datasets. Results are averaged over five random seeds under the unified protocol of \S\ref{sec:experiments}.}
\label{tab:main_results}
\centering
\setlength{\tabcolsep}{3.5pt}
\renewcommand{\arraystretch}{1.05}
\small
\resizebox{\linewidth}{!}{%
\begin{tabular}{@{}llrrrrrr@{}}
\toprule
Method & Metric & PAMAP2 & DSADS & MHealth & RealWorld & UCI-HAR & USC-HAD \\
\midrule
\multirow{2}{*}{BYOL + perm\_jit}
& Acc. & $84.06 \pm 8.26$ & $89.18 \pm 5.02$ & $85.49 \pm 7.66$ & $95.52 \pm 0.27$ & $77.65 \pm 3.80$ & $78.45 \pm 5.40$ \\
& F1   & $80.53 \pm 13.74$ & $88.73 \pm 5.45$ & $79.57 \pm 11.37$ & $95.60 \pm 0.40$ & $75.94 \pm 4.48$ & $73.43 \pm 6.08$ \\
\midrule
\multirow{2}{*}{BYOL + lfc}
& Acc. & $90.82 \pm 1.20$ & $96.71 \pm 0.14$ & $92.28 \pm 1.60$ & $96.11 \pm 0.34$ & $81.52 \pm 4.04$ & $82.85 \pm 1.20$ \\
& F1   & $90.66 \pm 1.45$ & $96.71 \pm 0.14$ & $91.41 \pm 1.76$ & $96.14 \pm 0.46$ & $80.43 \pm 4.76$ & $77.57 \pm 1.30$ \\
\midrule
\multirow{2}{*}{BYOL + Mixup}
& Acc. & $89.76 \pm 0.89$ & $97.00 \pm 0.52$ & $92.09 \pm 1.36$ & $96.00 \pm 0.19$ & $81.66 \pm 3.67$ & $81.28 \pm 2.44$ \\
& F1   & $89.53 \pm 1.07$ & $96.97 \pm 0.52$ & $91.75 \pm 1.52$ & $96.00 \pm 0.20$ & $80.54 \pm 4.56$ & $75.88 \pm 3.34$ \\
\midrule
\multirow{2}{*}{ModCL}
& Acc. & $90.60 \pm 1.17$ & $96.60 \pm 2.48$ & $94.51 \pm 4.09$ & $95.37 \pm 0.87$ & $89.20 \pm 3.84$ & $84.54 \pm 3.39$ \\
& F1   & $90.47 \pm 1.11$ & $96.57 \pm 2.54$ & $92.70 \pm 5.53$ & $95.42 \pm 0.87$ & $88.99 \pm 4.19$ & $81.06 \pm 3.08$ \\
\midrule
\multirow{2}{*}{TSLANet}
& Acc. & $92.20 \pm 0.58$ & $98.89 \pm 0.15$ & $97.12 \pm 1.29$ & $97.04 \pm 0.16$ & $85.58 \pm 8.30$ & $82.93 \pm 0.58$ \\
& F1   & $92.16 \pm 0.50$ & $98.87 \pm 0.16$ & $96.99 \pm 1.60$ & $97.11 \pm 0.18$ & $84.86 \pm 9.32$ & $84.28 \pm 0.60$ \\
\midrule
\multirow{2}{*}{CALANet}
& Acc. & $84.60 \pm 0.74$ & $94.91 \pm 0.11$ & $92.19 \pm 1.56$ & $93.66 \pm 0.35$ & $88.10 \pm 0.76$ & $77.61 \pm 1.49$ \\
& F1   & $81.56 \pm 1.00$ & $94.83 \pm 0.12$ & $90.66 \pm 1.92$ & $91.51 \pm 0.64$ & $87.56 \pm 0.85$ & $72.19 \pm 1.87$ \\
\midrule
\multirow{2}{*}{MoPFormer}
& Acc. & $90.64 \pm 0.52$ & $82.09 \pm 5.46$ & $97.15 \pm 1.20$ & $91.39 \pm 2.81$ & $83.37 \pm 7.18$ & $84.23 \pm 1.04$ \\
& F1   & $90.24 \pm 0.46$ & $79.44 \pm 6.58$ & $97.16 \pm 1.14$ & $92.13 \pm 2.32$ & $79.37 \pm 10.47$ & $80.64 \pm 1.11$ \\
\midrule

\multirow{2}{*}{HAR-JEPA}
& Acc. & $91.67 \pm 0.19$ & $97.88 \pm 0.05$ & $98.43 \pm 0.23$ & $95.24 \pm 0.10$ & $89.83 \pm 0.15$ & $86.56 \pm 0.21$ \\
& F1   & $91.96 \pm 0.20$ & $97.86 \pm 0.05$ & $98.54 \pm 0.21$ & $95.46 \pm 0.11$ & $90.14 \pm 0.14$ & $82.77 \pm 0.36$ \\
\midrule

\multirow{2}{*}{W-JEPA (no wavelets)}
  & Acc. & $91.46 \pm 0.39$ & $96.94 \pm 0.15$ & $96.00 \pm 0.38$ & $96.71 \pm 0.16$ & $88.64 \pm 1.26$ & $85.81 \pm 0.36$ \\
  & F1   & $91.29 \pm 0.34$ & $96.94 \pm 0.16$ & $95.55 \pm 0.21$ & $96.55 \pm 0.13$ & $87.80 \pm 1.28$ & $82.66 \pm 0.36$ \\
\midrule
\multirow{2}{*}{\textbf{W-JEPA}}
& Acc. & $\boldsymbol{93.75 \pm 0.22}$ & $\boldsymbol{99.22 \pm 0.13}$ & $\boldsymbol{99.28 \pm 0.35}$
& $\boldsymbol{97.13 \pm 0.14}$ & $\boldsymbol{92.91 \pm 1.23}$ & $\boldsymbol{88.54 \pm 0.35}$ \\
& F1   & $\boldsymbol{93.79 \pm 0.21}$ & $\boldsymbol{99.22 \pm 0.13}$ & $\boldsymbol{99.18 \pm 0.19}$
& $\boldsymbol{97.36 \pm 0.12}$ & $\boldsymbol{92.59 \pm 1.26}$ & $\boldsymbol{86.52 \pm 0.33}$ \\
\bottomrule
\end{tabular}%
}
\end{table}

\subsection{What the wavelet tokens contribute}
\label{sec:tokenizer_analysis}

\paragraph{Replacing wavelet tokens with raw patches.} The variant \emph{W-JEPA (no wavelets)} replaces the wavelet tokenizer
with a conventional temporal-patch tokenizer while holding the remainder
of the pipeline fixed. Each 500-sample channel is divided into ten
consecutive 50-sample patches; the ordered, signed samples of a patch are
embedded by an MLP of the same depth and width. With wavelet tokens, subject-disjoint macro-F1 is higher on all six datasets (Table~\ref{tab:subject_results}). Two JEPAs that predict temporal-patch units, HAR-JEPA and W-JEPA (no wavelets), reach 71.9 and 73.1 average held-out macro-F1; replacing the patches with wavelet tokens raises W-JEPA to 80.6. This comparison supports the tokenizer as a useful design choice within W-JEPA.

\begin{figure}[ht]
    \centering
    \includegraphics[width=0.90\textwidth]{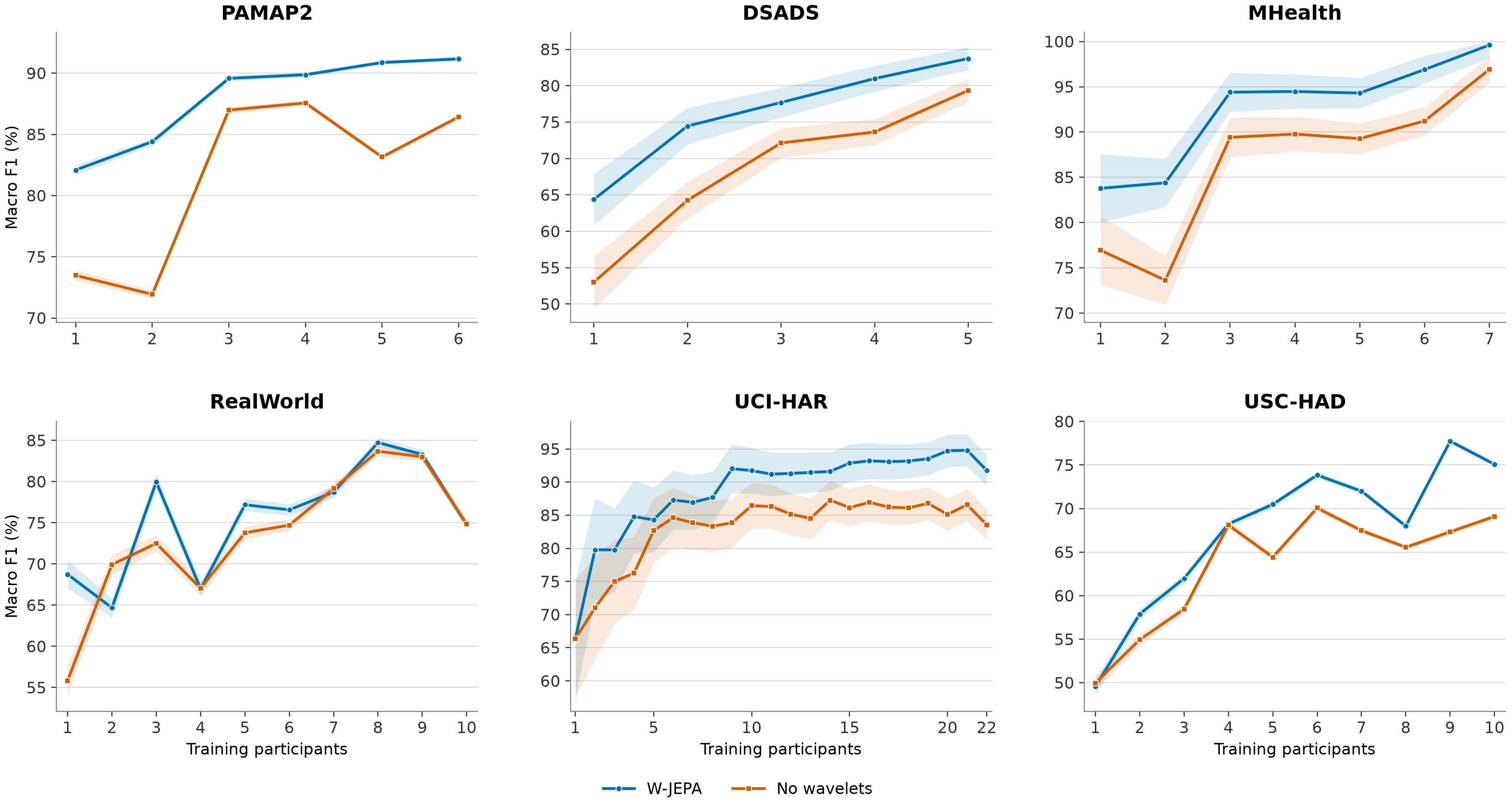}
    \caption{Subject-disjoint macro-F1 with wavelet tokens and learned temporal
    patches as training participants are added. A fixed 20\% participant split
    is held out for testing, and training participants are drawn from the
    remaining pool.}
    \label{fig:subject_disjoint_scaling}
\end{figure}

\paragraph{Varying participant diversity.}
Figure~\ref{fig:subject_disjoint_scaling} compares the tokenizers with 20\% of
participants held out for testing. This test cohort remains fixed as complete
participants from the remaining pool are progressively added to the
fine-tuning set. At matched training set sizes, W-JEPA generally outperforms
the \emph{no wavelets} variant, especially when few participants are available.
The advantage persists across the displayed training-set sizes on PAMAP2,
DSADS, and MHealth, while RealWorld shows smaller and less consistent gains.
With the test cohort fixed, the curves compare the benefit of adding training
participants on the same evaluation population. They provide evidence that
wavelet tokens improve performance under limited participant diversity.

\begin{figure}[ht]
    \centering
    \begin{minipage}[t]{0.49\linewidth}
        \centering
        \includegraphics[width=\linewidth]{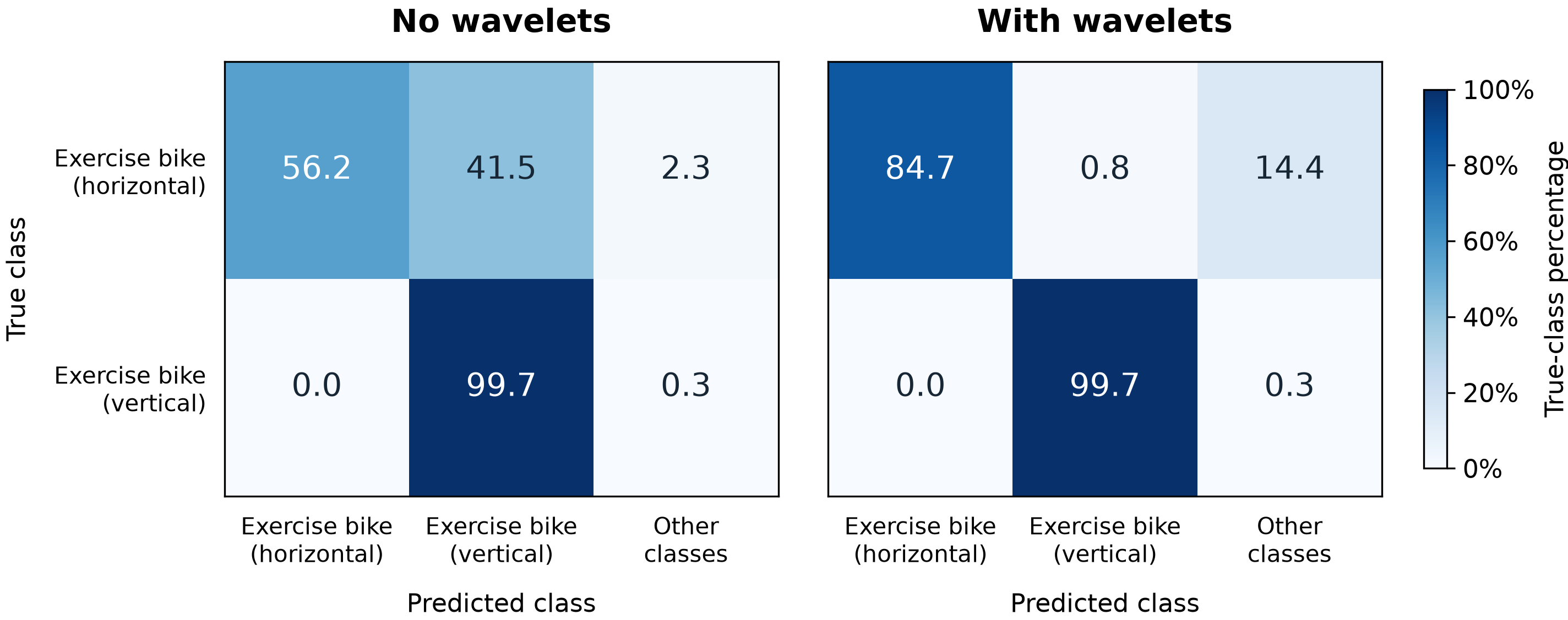}
        \par\smallskip
        {\small (a) Confusion between cycling activities}
    \end{minipage}\hfill
    \begin{minipage}[t]{0.49\linewidth}
        \centering
        \includegraphics[width=\linewidth]{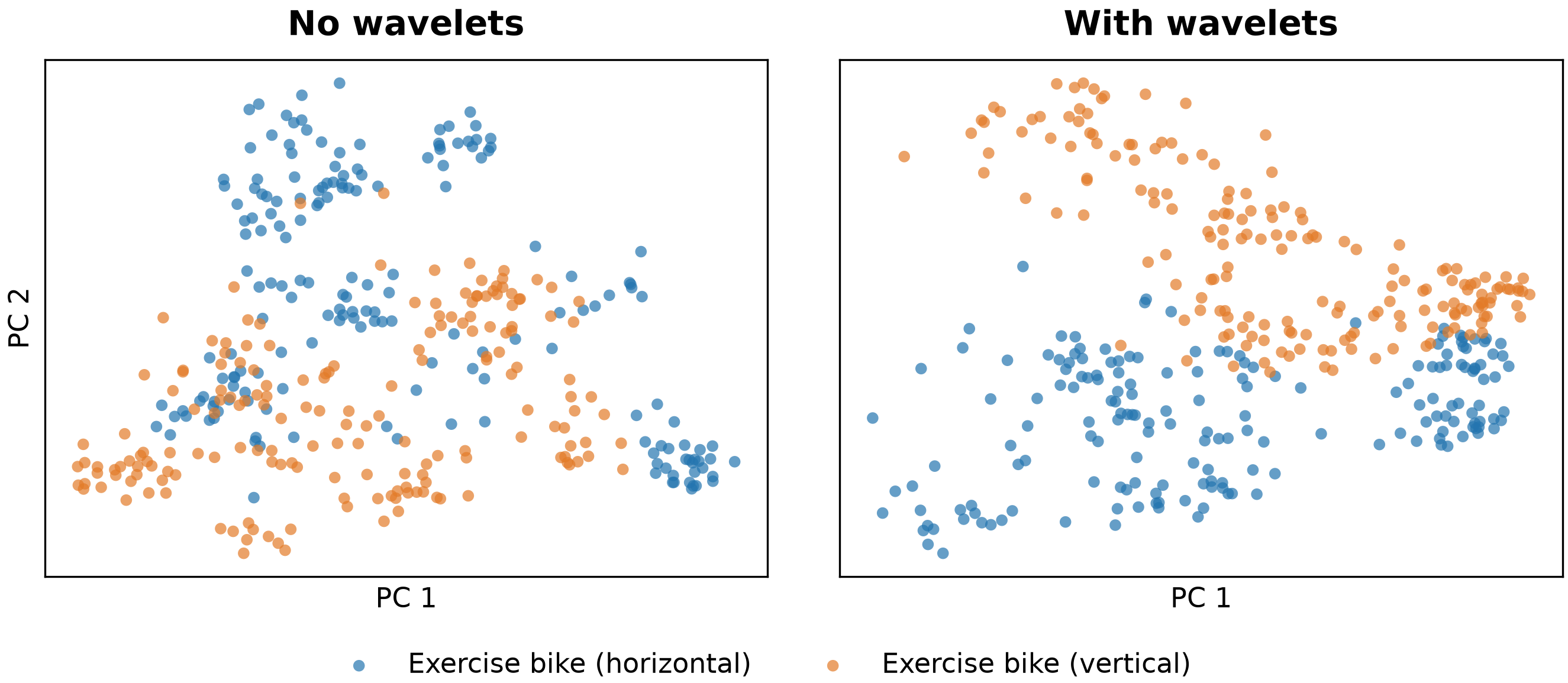}
        \par\smallskip
        {\small (b) PCA of learned representations}
    \end{minipage}
    \caption{Each panel compares W-JEPA
    without and with wavelets. (a) Confusion matrices for cycling on a horizontal
    exercise bike and cycling on a vertical exercise bike, using percentages within each true class; predictions of other
    activities are ``Other classes.'' (b) 2D PCA
    projections of learned representations for the same two classes.}
    \label{fig:dsads_cycling}
\end{figure}

\paragraph{Separating near-identical activities.}
\label{sec:dsads_cycling}
We return to the example of the exercise-bike pair of Figure~\ref{fig:activity_detail}, where we further compare W-JEPA with and without wavelet tokenization on DSADS under the same subject-disjoint evaluation protocol, keeping pretraining, fine-tuning, and all other experimental settings fixed. (Figure~\ref{fig:dsads_cycling}). Without wavelets, 41.5\% of horizontal-bike
windows are classified as vertical-bike cycling; with wavelets, this falls to
0.8\%, improving horizontal-bike recall while preserving vertical-bike recall; some errors shift to other activities. In the PCA projection of the
learned representations, the two classes overlap without wavelets and are more separated
with them. This is consistent with wavelet tokens exposing scale-specific differences in
Figure~\ref{fig:activity_detail}.

\paragraph{Interpreting attention by timescale.}
Because every token carries a scale address, the fine-tuned encoder's attention
can be read by timescale, which a temporal-patch tokenizer does not allow. We
measure the attention that tokens receive in the final encoder layer, averaged
over heads and query positions and summed within each band, on held-out DSADS
participants (Appendix~\ref{sec:attention_noise},
Figure~\ref{fig:scale_attention}). On clean inputs, the coarse $A_4$ band receives the largest share for most
activities, whereas, $D_1$, receives the largest share for
running and jumping. We then add noise at increasing levels, measured against
each channel's training RMS. Attention shifts from $A_4$ toward the detail
bands: at 30\% noise, $A_4$ attention falls by 6.1\% relative to clean inputs,
while $D_2$ and $D_1$ rise by 11.1\% and 9.4\%. Over the same range, held-out
accuracy and macro-F1 stay within about one point of their clean values
(Appendix Table~\ref{tab:dsads_noise_performance}).

\subsection{Ablation under restricted sensor settings}
\label{sec:restricted_settings}

We evaluate a restricted setting in which downstream fine-tuning uses only
wrist accelerometer and gyroscope channels: nine channels for PAMAP2 and six
for MHealth. Pretraining follows the same protocol as in
\S\ref{sec:experiments}, using all datasets in the pretraining pool except the
evaluation dataset. All other experimental settings are kept the same.
We evaluate both random-window and subject-disjoint splits, using 20\% of
windows or participants, respectively, for training and the remaining 80\%
for evaluation. Each experiment uses five training seeds. Tables ~\ref{tab:restricted_random} and ~\ref{tab:restricted_subject} report
macro-F1 and accuracy under these restrictions. W-JEPA achieves the highest
mean score on both metrics for both datasets under both split protocols. Scores are lower with participant holdout for every method, while
W-JEPA retains its lead in mean performance when fine-tuning is restricted
to wrist inertial signals.

\begin{table}[ht]
\captionsetup{justification=centering,singlelinecheck=false}
\caption{Restricted sensor ablation with random-window splits.}
\label{tab:restricted_random}
\centering
\footnotesize
\setlength{\tabcolsep}{2.5pt}
\renewcommand{\arraystretch}{0.95}
\begin{adjustbox}{max width=\linewidth}
\begin{tabular}{@{}lrrrr@{}}
\toprule
\multirow{2}{*}{Method}
& \multicolumn{2}{c}{PAMAP2}
& \multicolumn{2}{c}{MHealth} \\
\cmidrule(lr){2-3}\cmidrule(l){4-5}
& \multicolumn{1}{c}{F1}
& \multicolumn{1}{c}{Acc.}
& \multicolumn{1}{c}{F1}
& \multicolumn{1}{c}{Acc.} \\
\midrule
BYOL + perm\_jit
& $58.97 \pm 6.41$ & $67.32 \pm 3.75$
& $40.77 \pm 11.21$ & $48.50 \pm 6.07$ \\
BYOL + lfc
& $77.85 \pm 1.13$ & $79.60 \pm 0.71$
& $49.63 \pm 3.19$ & $55.08 \pm 2.77$ \\
BYOL + Mixup
& $75.16 \pm 3.44$ & $78.28 \pm 2.03$
& $47.31 \pm 3.06$ & $52.76 \pm 4.03$ \\
ModCL
& $75.99 \pm 5.84$ & $79.10 \pm 3.63$
& $72.84 \pm 4.67$ & $78.79 \pm 2.21$ \\
TSLANet
& $87.63 \pm 0.35$ & $86.79 \pm 0.56$
& $77.54 \pm 2.29$ & $78.02 \pm 2.57$ \\
CALANet
& $74.63 \pm 1.14$ & $78.11 \pm 0.58$
& $75.50 \pm 1.17$ & $77.60 \pm 1.20$ \\
MoPFormer
& $85.14 \pm 0.38$ & $85.52 \pm 0.30$
& $88.81 \pm 0.67$ & $88.95 \pm 0.83$ \\
HAR-JEPA
& $85.96 \pm 0.26$ & $86.31 \pm 0.28$
& $90.91 \pm 0.88$ & $90.67 \pm 0.88$ \\
\midrule
W-JEPA (no wavelets)
& $86.59 \pm 0.31$ & $86.54 \pm 0.28$
& $89.58 \pm 0.55$ & $90.47 \pm 0.63$ \\
\textbf{W-JEPA}
& $\boldsymbol{88.10 \pm 0.11}$
& $\boldsymbol{88.09 \pm 0.14}$
& $\boldsymbol{93.53 \pm 0.54}$
& $\boldsymbol{94.22 \pm 0.76}$ \\
\bottomrule
\end{tabular}
\end{adjustbox}
\end{table}

\begin{table}[ht]
\captionsetup{justification=centering,singlelinecheck=false}
\caption{Restricted sensor ablation with subject-disjoint splits.}
\label{tab:restricted_subject}
\centering
\footnotesize
\setlength{\tabcolsep}{2.5pt}
\renewcommand{\arraystretch}{0.95}
\begin{adjustbox}{max width=\linewidth}
\begin{tabular}{@{}lrrrr@{}}
\toprule
\multirow{2}{*}{Method}
& \multicolumn{2}{c}{PAMAP2}
& \multicolumn{2}{c}{MHealth} \\
\cmidrule(lr){2-3}\cmidrule(l){4-5}
& \multicolumn{1}{c}{F1}
& \multicolumn{1}{c}{Acc.}
& \multicolumn{1}{c}{F1}
& \multicolumn{1}{c}{Acc.} \\
\midrule
BYOL + perm\_jit
& $46.88 \pm 1.94$ & $50.23 \pm 4.10$
& $35.01 \pm 6.01$ & $40.87 \pm 3.42$ \\
BYOL + lfc
& $48.57 \pm 1.54$ & $47.98 \pm 2.22$
& $39.08 \pm 1.10$ & $43.42 \pm 2.30$ \\
BYOL + Mixup
& $51.88 \pm 1.64$ & $52.06 \pm 2.46$
& $40.23 \pm 1.19$ & $44.08 \pm 2.77$ \\
ModCL
& $61.71 \pm 1.66$ & $62.63 \pm 2.44$
& $51.26 \pm 6.29$ & $53.89 \pm 5.05$ \\
TSLANet
& $63.07 \pm 3.29$ & $61.98 \pm 3.60$
& $59.24 \pm 2.97$ & $58.43 \pm 2.74$ \\
CALANet
& $55.90 \pm 1.75$ & $55.14 \pm 1.63$
& $59.44 \pm 1.73$ & $61.63 \pm 1.58$ \\
MoPFormer
& $58.71 \pm 0.51$ & $59.57 \pm 1.16$
& $66.17 \pm 3.26$ & $68.19 \pm 2.08$ \\
HAR-JEPA
& $60.54 \pm 0.26$ & $61.57 \pm 0.32$
& $65.29 \pm 0.96$ & $66.39 \pm 1.09$ \\
\midrule
W-JEPA (no wavelets)
& $61.93 \pm 1.64$ & $62.16 \pm 1.77$
& $64.26 \pm 1.79$ & $64.77 \pm 0.92$ \\
\textbf{W-JEPA}
& $\boldsymbol{64.14 \pm 2.12}$
& $\boldsymbol{64.25 \pm 1.16}$
& $\boldsymbol{67.63 \pm 1.13}$
& $\boldsymbol{69.45 \pm 2.62}$ \\
\bottomrule
\end{tabular}
\end{adjustbox}
\end{table}

\section{Conclusion}
\label{sec:conclusion}

We asked how the units of masked latent prediction should be defined for
heterogeneous wearable data. W-JEPA organizes inertial windows into wavelet tokens indexed by scale, sensor channel axis, and time, and learns by predicting masked latent features. It outperforms the evaluated baselines, and loses the least when evaluating on unseen participants. Replacing wavelet tokens with temporal patches weakens performance and can increase confusion between similar activities. These
results suggest that physically grounded multiscale tokenization and latent
prediction together provide a promising foundation for transferable
representations of heterogeneous wearable data.

\subsubsection*{Acknowledgments}
This work was supported by funding from the National Heart, Lung, and Blood Institute (NIH/NHLBI 1R01HL166802 and 1R01HL168535).

\clearpage
\bibliography{iclr2027_conference}

@article{chen2021deep,
  title={Deep learning for sensor-based human activity recognition: Overview, challenges, and opportunities},
  author={Chen, Kaixuan and Zhang, Dalin and Yao, Lina and Guo, Bin and Yu, Zhiwen and Liu, Yunhao},
  journal={ACM Computing Surveys (CSUR)},
  volume={54},
  number={4},
  pages={1--40},
  year={2021},
  publisher={ACM New York, NY, USA}
}

@inproceedings{xu2021limubert,
author = {Xu, Huatao and Zhou, Pengfei and Tan, Rui and Li, Mo and Shen, Guobin},
title = {LIMU-BERT: Unleashing the Potential of Unlabeled Data for IMU Sensing Applications},
year = {2021},
isbn = {9781450390972},
publisher = {Association for Computing Machinery},
address = {New York, NY, USA},
url = {https://doi.org/10.1145/3485730.3485937},
doi = {10.1145/3485730.3485937},
booktitle = {Proceedings of the 19th ACM Conference on Embedded Networked Sensor Systems},
pages = {220–233},
numpages = {14},
location = {Coimbra, Portugal},
series = {SenSys '21}
}

@misc{lu2023diversify,
      title={Out-of-Distribution Representation Learning for Time Series Classification}, 
      author={Wang Lu and Jindong Wang and Xinwei Sun and Yiqiang Chen and Xing Xie},
      year={2023},
      eprint={2209.07027},
      archivePrefix={arXiv},
      primaryClass={cs.LG},
      url={https://arxiv.org/abs/2209.07027}, 
}

@inproceedings{zhang2024unimts,
 author = {Zhang, Xiyuan and Teng, Diyan and Chowdhury, Ranak Roy and Li, Shuheng and Hong, Dezhi and Gupta, Rajesh K. and Shang, Jingbo},
 booktitle = {Advances in Neural Information Processing Systems},
 doi = {10.52202/079017-3414},
 editor = {A. Globerson and L. Mackey and D. Belgrave and A. Fan and U. Paquet and J. Tomczak and C. Zhang},
 pages = {107469--107493},
 publisher = {Curran Associates, Inc.},
 title = {UniMTS: Unified Pre-training for Motion Time Series},
 url = {https://proceedings.neurips.cc/paper_files/paper/2024/file/c290d4373c495b2cad0625d6288260f0-Paper-Conference.pdf},
 volume = {37},
 year = {2024}
}

@misc{ding2026halo,
      title={HALO: A Heterogeneity-Aware Language-Aligned IMU Foundation Model for Open-Set Human Activity Recognition}, 
      author={Zihan Ding and Liyu Zhang and Xiaomin Ouyang},
      year={2026},
      eprint={2608.27233},
      archivePrefix={arXiv},
      primaryClass={cs.LG},
      url={https://arxiv.org/abs/2608.27233}, 
}

@misc{thukral2026wavelet,
      title={Wavelet-Driven Masked Multiscale Reconstruction for PPG Foundation Models}, 
      author={Megha Thukral and Cyrus Tanade and Simon A. Lee and Juhyeon Lee and Hao Zhou and Keum San Chun and Migyeong Gwak and Viswam Nathan and Md Mahbubur Rahman and Li Zhu and Mehrab Bin Morshed and Subramaniam Venkatraman and Sharanya Arcot Desai},
      year={2026},
      eprint={2601.12215},
      archivePrefix={arXiv},
      primaryClass={cs.LG},
      url={https://arxiv.org/abs/2601.12215}, 
}

@inproceedings{hammerla2015lets,
author = {Hammerla, Nils Y. and Pl{\"o}tz, Thomas},
title = {Let's (not) stick together: pairwise similarity biases cross-validation in activity recognition},
year = {2015},
isbn = {9781450335744},
publisher = {Association for Computing Machinery},
address = {New York, NY, USA},
url = {https://doi.org/10.1145/2750858.2807551},
doi = {10.1145/2750858.2807551},
booktitle = {Proceedings of the 2015 ACM International Joint Conference on Pervasive and Ubiquitous Computing},
pages = {1041–1051},
numpages = {11},
location = {Osaka, Japan},
series = {UbiComp '15}
}

@article{hong2024crosshar,
author = {Hong, Zhiqing and Li, Zelong and Zhong, Shuxin and Lyu, Wenjun and Wang, Haotian and Ding, Yi and He, Tian and Zhang, Desheng},
title = {CrossHAR: Generalizing Cross-dataset Human Activity Recognition via Hierarchical Self-Supervised Pretraining},
year = {2024},
issue_date = {June 2024},
publisher = {Association for Computing Machinery},
address = {New York, NY, USA},
volume = {8},
number = {2},
url = {https://doi.org/10.1145/3659597},
doi = {10.1145/3659597},
journal = {Proc. ACM Interact. Mob. Wearable Ubiquitous Technol.},
month = may,
articleno = {64},
numpages = {26}
}

@inproceedings{xu2023unihar,
author = {Xu, Huatao and Zhou, Pengfei and Tan, Rui and Li, Mo},
title = {Practically Adopting Human Activity Recognition},
year = {2023},
isbn = {9781450399906},
publisher = {Association for Computing Machinery},
address = {New York, NY, USA},
url = {https://doi.org/10.1145/3570361.3613299},
doi = {10.1145/3570361.3613299},
booktitle = {Proceedings of the 29th Annual International Conference on Mobile Computing and Networking},
articleno = {85},
numpages = {15},
location = {Madrid, Spain},
series = {ACM MobiCom '23}
}

@article{ryu2025moca,
  title={Moca: Multi-modal cross-masked autoencoder for digital health measurements},
  author={Ryu, Howon and Chen, Yuliang and Wang, Yacun and LaCroix, Andrea Z and Di, Chongzhi and Natarajan, Loki and Wang, Yu and Zou, Jingjing},
  journal={arXiv preprint arXiv:2506.02260},
  year={2025}
}

@inproceedings{haresamudram2020masked,
author = {Haresamudram, Harish and Beedu, Apoorva and Agrawal, Varun and Grady, Patrick L. and Essa, Irfan and Hoffman, Judy and Pl{\"o}tz, Thomas},
title = {Masked reconstruction based self-supervision for human activity recognition},
year = {2020},
isbn = {9781450380775},
publisher = {Association for Computing Machinery},
address = {New York, NY, USA},
url = {https://doi.org/10.1145/3410531.3414306},
doi = {10.1145/3410531.3414306},
booktitle = {Proceedings of the 2020 ACM International Symposium on Wearable Computers},
pages = {45–49},
numpages = {5},
location = {Virtual Event, Mexico},
series = {ISWC '20}
}

@misc{noor2026joint,
      title={Joint-Embedding Predictive Architecture for Sensor-based Activity Recognition}, 
      author={Mohd Halim Mohd Noor and Abdulrahman M. A. Baraka},
      year={2026},
      eprint={2607.16350},
      archivePrefix={arXiv},
      primaryClass={eess.SP},
      url={https://arxiv.org/abs/2607.16350}, 
}

@article{grill2020bootstrap,
  title={Bootstrap your own latent-a new approach to self-supervised learning},
  author={Grill, Jean-Bastien and Strub, Florian and Altch{\'e}, Florent and Tallec, Corentin and Richemond, Pierre and Buchatskaya, Elena and Doersch, Carl and Avila Pires, Bernardo and Guo, Zhaohan and Gheshlaghi Azar, Mohammad and others},
  journal={Advances in neural information processing systems},
  volume={33},
  pages={21271--21284},
  year={2020}
}

@inproceedings{qian2022makes,
author = {Qian, Hangwei and Tian, Tian and Miao, Chunyan},
title = {What Makes Good Contrastive Learning on Small-Scale Wearable-based Tasks?},
year = {2022},
isbn = {9781450393850},
publisher = {Association for Computing Machinery},
address = {New York, NY, USA},
url = {https://doi.org/10.1145/3534678.3539134},
doi = {10.1145/3534678.3539134},
booktitle = {Proceedings of the 28th ACM SIGKDD Conference on Knowledge Discovery and Data Mining},
pages = {3761--3771},
numpages = {11},
location = {Washington DC, USA},
series = {KDD '22}
}

@ARTICLE{guo2024modality,
  author={Guo, Changru and Zhang, Yingwei and Chen, Yiqiang and Xu, Chenyang and Wang, Zhong},
  journal={IEEE Internet of Things Journal}, 
  title={Modality Consistency-Guided Contrastive Learning for Wearable-Based Human Activity Recognition}, 
  year={2024},
  volume={11},
  number={12},
  pages={21750--21762},
  doi={10.1109/JIOT.2024.3379019}}

@inproceedings{eldele2024tslanet,
author = {Eldele, Emadeldeen and Ragab, Mohamed and Chen, Zhenghua and Wu, Min and Li, Xiaoli},
title = {{TSLANet}: rethinking transformers for time series representation learning},
year = {2024},
publisher = {JMLR.org},
booktitle = {Proceedings of the 41st International Conference on Machine Learning},
articleno = {494},
numpages = {20},
location = {Vienna, Austria},
series = {ICML'24}
}

@inproceedings{park2024calanet,
 author = {Park, Jaegyun and Kim, Dae-Won and Lee, Jaesung},
 booktitle = {Advances in Neural Information Processing Systems},
 editor = {A. Globerson and L. Mackey and D. Belgrave and A. Fan and U. Paquet and J. Tomczak and C. Zhang},
 pages = {69419--69444},
 publisher = {Curran Associates, Inc.},
 title = {{CALANet}: Cheap All-Layer Aggregation for Human Activity Recognition},
 url = {https://proceedings.neurips.cc/paper_files/paper/2024/file/8078e76f913e31b8467e85b4c0f0d22b-Paper-Conference.pdf},
 volume = {37},
 year = {2024}
}

@inproceedings{zhang2026mopformer,
author = {Zhang, Hao and Zhuang, Zhan and Wang, Xuehao and Yang, Xiaodong and Zhang, Yu},
 booktitle = {Advances in Neural Information Processing Systems},
 doi = {10.52202/085713-1313},
 editor = {D. Belgrave and C. Zhang and H. Lin and R. Pascanu and P. Koniusz and M. Ghassemi and N. Chen},
 pages = {39334--39360},
 publisher = {Curran Associates, Inc.},
 title = {MoPFormer: Motion-Primitive Transformer for Wearable-Sensor Activity Recognition},
 url = {https://proceedings.neurips.cc/paper_files/paper/2025/file/37ecb317d11004f77defbca54ef3b928-Paper-Conference.pdf},
 volume = {38, Main Conference},
 year = {2025}
}

@inproceedings{zerveas2021transformer,
  title={A transformer-based framework for multivariate time series representation learning},
  author={Zerveas, George and Jayaraman, Srideepika and Patel, Dhaval and Bhamidipaty, Anuradha and Eickhoff, Carsten},
  booktitle={Proceedings of the 27th ACM SIGKDD conference on knowledge discovery \& data mining},
  pages={2114--2124},
  year={2021}
}

@article{eldele2021time,
  title={Time-series representation learning via temporal and contextual contrasting},
  author={Eldele, Emadeldeen and Ragab, Mohamed and Chen, Zhenghua and Wu, Min and Kwoh, Chee Keong and Li, Xiaoli and Guan, Cuntai},
  journal={arXiv preprint arXiv:2106.14112},
  year={2021}
}

@article{zhang2022tfc,
  title={Self-supervised contrastive pre-training for time series via time-frequency consistency},
  author={Zhang, Xiang and Zhao, Ziyuan and Tsiligkaridis, Theodoros and Zitnik, Marinka},
  journal={Advances in neural information processing systems},
  volume={35},
  pages={3988--4003},
  year={2022}
}

@article{balestriero2025lejepa,
  title={Lejepa: Provable and scalable self-supervised learning without the heuristics},
  author={Balestriero, Randall and LeCun, Yann},
  journal={arXiv preprint arXiv:2511.08544},
  year={2025}
}

@article{lecun2022path,
  title={A path towards autonomous machine intelligence version 0.9. 2, 2022-06-27},
  author={LeCun, Yann and others},
  journal={Open Review},
  volume={62},
  number={1},
  pages={1--62},
  year={2022}
}

@inproceedings{assran2023self,
  title={Self-supervised learning from images with a joint-embedding predictive architecture},
  author={Assran, Mahmoud and Duval, Quentin and Misra, Ishan and Bojanowski, Piotr and Vincent, Pascal and Rabbat, Michael and LeCun, Yann and Ballas, Nicolas},
  booktitle={2023 IEEE/CVF Conference on Computer Vision and Pattern Recognition (CVPR)},
  pages={15619--15629},
  year={2023},
  organization={IEEE}
}

@article{bardes2024revisiting,
  title={Revisiting feature prediction for learning visual representations from video},
  author={Bardes, Adrien and Garrido, Quentin and Ponce, Jean and Chen, Xinlei and Rabbat, Michael and LeCun, Yann and Assran, Mahmoud and Ballas, Nicolas},
  journal={arXiv preprint arXiv:2404.08471},
  year={2024}
}

@article{ennadir2025joint,
  title={Joint embeddings go temporal},
  author={Ennadir, Sofiane and Golkar, Siavash and Sarra, Leopoldo},
  journal={arXiv preprint arXiv:2509.25449},
  year={2025}
}

@inproceedings{abdelfattah2024s,
  title={S-jepa: A joint embedding predictive architecture for skeletal action recognition},
  author={Abdelfattah, Mohamed and Alahi, Alexandre},
  booktitle={European Conference on Computer Vision},
  pages={367--384},
  year={2024},
  organization={Springer}
}

@article{guetschel2024s,
  title={S-JEPA: towards seamless cross-dataset transfer through dynamic spatial attention},
  author={Guetschel, Pierre and Moreau, Thomas and Tangermann, Michael},
  journal={arXiv preprint arXiv:2403.11772},
  year={2024}
}

@book{daubechies1992ten,
  title={Ten lectures on wavelets},
  author={Daubechies, Ingrid},
  year={1992},
  publisher={SIAM}
}

@article{tian2020wavelet,
  title={Robust human activity recognition using single accelerometer via wavelet energy spectrum features and ensemble feature selection},
  author={Tian, Yiming and Zhang, Jie and Wang, Jie and Geng, Yanli and Wang, Xitai},
  journal={Systems Science \& Control Engineering},
  volume={8},
  number={1},
  pages={83--96},
  year={2020},
  publisher={Taylor \& Francis}
}

@article{vuong2023wavelet,
  title={Deep wavelet convolutional neural networks for multimodal human activity recognition using wearable inertial sensors},
  author={Vuong, Thi Hong and Doan, Tung and Takasu, Atsuhiro},
  journal={Sensors},
  volume={23},
  number={24},
  pages={9721},
  year={2023},
  publisher={MDPI}
}

@article{masserano2024enhancing,
  title={Enhancing foundation models for time series forecasting via wavelet-based tokenization},
  author={Masserano, Luca and Ansari, Abdul Fatir and Han, Boran and Zhang, Xiyuan and Faloutsos, Christos and Mahoney, Michael W and Wilson, Andrew Gordon and Park, Youngsuk and Rangapuram, Syama and Maddix, Danielle C and others},
  journal={arXiv preprint arXiv:2412.05244},
  year={2024}
}

@article{wolter2024ptwt,
  title={ptwt-the PyTorch wavelet toolbox},
  author={Wolter, Moritz and Blanke, Felix and Garcke, Jochen and Hoyt, Charles Tapley},
  journal={Journal of Machine Learning Research},
  volume={25},
  number={80},
  pages={1--7},
  year={2024}
}

@article{bardes2022vicreg,
  title={Vicreg: Variance-invariance-covariance regularization for self-supervised learning},
  author={Bardes, Adrien and Ponce, Jean and LeCun, Yann},
  journal={arXiv preprint arXiv:2105.04906},
  year={2021}
}

@misc{loshchilov2019adamw,
      title={Decoupled Weight Decay Regularization}, 
      author={Ilya Loshchilov and Frank Hutter},
      year={2019},
      eprint={1711.05101},
      archivePrefix={arXiv},
      primaryClass={cs.LG},
      url={https://arxiv.org/abs/1711.05101}, 
}

@inproceedings{reiss2012pamap2,
  author={Reiss, Attila and Stricker, Didier},
  booktitle={2012 16th International Symposium on Wearable Computers}, 
  title={Introducing a New Benchmarked Dataset for Activity Monitoring}, 
  year={2012},
  volume={},
  number={},
  pages={108-109},
  doi={10.1109/ISWC.2012.13}}

@article{altun2010dsads,
title = {Comparative study on classifying human activities with miniature inertial and magnetic sensors},
journal = {Pattern Recognition},
volume = {43},
number = {10},
pages = {3605-3620},
year = {2010},
issn = {0031-3203},
doi = {https://doi.org/10.1016/j.patcog.2010.04.019},
url = {https://www.sciencedirect.com/science/article/pii/S0031320310001950},
author = {Kerem Altun and Billur Barshan and Orkun Tunçel},
}

@misc{banos2014mhealth,
  author = {Banos, Oresti and Garcia, Rafael and Saez, Alejandro},
  title        = {{MHEALTH}},
  year         = {2014},
  howpublished = {UCI Machine Learning Repository},
  note         = {{DOI}: https://doi.org/10.24432/C5TW22}
}

@inproceedings{sztyler2016realworld,
  author={Sztyler, Timo and Stuckenschmidt, Heiner},
  booktitle={2016 IEEE International Conference on Pervasive Computing and Communications (PerCom)}, 
  title={On-body localization of wearable devices: An investigation of position-aware activity recognition}, 
  year={2016},
  volume={},
  number={},
  pages={1-9},
  doi={10.1109/PERCOM.2016.7456521}}

@misc{anguita2013ucihar,
  author       = {Reyes-Ortiz, Jorge Luis and Anguita, Davide and Ghio, Alessandro and Oneto, Luca and Parra, Xavier},
  title        = {{Human Activity Recognition Using Smartphones}},
  year         = {2013},
  howpublished = {UCI Machine Learning Repository},
  note         = {{DOI}: https://doi.org/10.24432/C54S4K}
}

@Article{s21196393,
AUTHOR = {Kim, Hyejoo and Kim, Hyeon-Joo and Park, Jinyoon and Ryu, Jeh-Kwang and Kim, Seung-Chan},
TITLE = {Recognition of Fine-Grained Walking Patterns Using a Smartwatch with Deep Attentive Neural Networks},
JOURNAL = {Sensors},
VOLUME = {21},
YEAR = {2021},
NUMBER = {19},
ARTICLE-NUMBER = {6393},
URL = {https://www.mdpi.com/1424-8220/21/19/6393},
PubMedID = {34640712},
ISSN = {1424-8220},
DOI = {10.3390/s21196393}
}

@inproceedings{zhang2012uschad,
author = {Zhang, Mi and Sawchuk, Alexander A.},
title = {USC-HAD: a daily activity dataset for ubiquitous activity recognition using wearable sensors},
year = {2012},
isbn = {9781450312240},
publisher = {Association for Computing Machinery},
address = {New York, NY, USA},
url = {https://doi.org/10.1145/2370216.2370438},
doi = {10.1145/2370216.2370438},
booktitle = {Proceedings of the 2012 ACM Conference on Ubiquitous Computing},
pages = {1036–1043},
numpages = {8},
location = {Pittsburgh, Pennsylvania},
series = {UbiComp '12}
}

@misc{pamap2_physical_activity_monitoring_231,
  author       = {Reiss, Attila},
  title        = {{PAMAP2 Physical Activity Monitoring}},
  year         = {2012},
  howpublished = {UCI Machine Learning Repository},
  note         = {{DOI}: https://doi.org/10.24432/C5NW2H}
}

@misc{daily_and_sports_activities_256,
  author       = {Barshan, Billur and Altun, Kerem},
  title        = {{Daily and Sports Activities}},
  year         = {2010},
  howpublished = {UCI Machine Learning Repository},
  note         = {{DOI}: https://doi.org/10.24432/C5C59F}
}

@article{banos2015design,
  title={Design, implementation and validation of a novel open framework for agile development of mobile health applications},
  author={Banos, Oresti and Villalonga, Claudia and Garcia, Rafael and Saez, Alejandro and Damas, Miguel and Holgado-Terriza, Juan A and Lee, Sungyong and Pomares, Hector and Rojas, Ignacio},
  journal={Biomedical engineering online},
  volume={14},
  pages={1--20},
  year={2015},
  publisher={Springer}
}

@inproceedings{Anguita2013APD,
  title={A Public Domain Dataset for Human Activity Recognition using Smartphones},
  author={D. Anguita and Alessandro Ghio and L. Oneto and Xavier Parra and Jorge Luis Reyes-Ortiz},
  booktitle={The European Symposium on Artificial Neural Networks},
  year={2013},
  url={https://api.semanticscholar.org/CorpusID:6975432}
}

@ARTICLE{10177231,
  author={Eldele, Emadeldeen and Ragab, Mohamed and Chen, Zhenghua and Wu, Min and Kwoh, Chee-Keong and Li, Xiaoli},
  journal={IEEE Transactions on Artificial Intelligence}, 
  title={Contrastive Domain Adaptation for Time-Series Via Temporal Mixup}, 
  year={2024},
  volume={5},
  number={3},
  pages={1185--1194},
  doi={10.1109/TAI.2023.3293473}}
\bibliographystyle{iclr2027_conference}


\clearpage
\appendix
\section{Datasets and Splits}
\label{sec:dataset_details}

\begin{table}[ht]
\caption{Dataset statistics for the subject-disjoint train/test splits. The number of
samples corresponds to the number of segmented sensor windows used for training
and testing.}
\label{tab:dataset_statistics}
\centering
\setlength{\tabcolsep}{6pt}
\renewcommand{\arraystretch}{1.05}
\small
\begin{tabular}{@{}lrrrr@{}}
\toprule
Dataset &
Train samples &
Test samples &
Train participants &
Test participants \\
\midrule
PAMAP2    & 835   & 2,998  & 2 & 7  \\
DSADS     & 2,242 & 6,726  & 2 & 6  \\
MHealth   & 267   & 1,068  & 2 & 8  \\
RealWorld & 2,551 & 10,209 & 3 & 12 \\
UCI-HAR   & 486   & 1,946  & 6 & 24 \\
USC-HAD   & 1,069 & 4,284  & 3 & 11 \\
\bottomrule
\end{tabular}
\end{table}

\paragraph{PAMAP2.}
The PAMAP2 Physical Activity Monitoring dataset contains recordings from nine
participants wearing inertial measurement units on the dominant wrist, chest,
and dominant ankle, together with a heart-rate monitor
\citep{pamap2_physical_activity_monitoring_231,reiss2012pamap2}. The inertial
units record acceleration, angular velocity, and magnetic field at 100 Hz,
while heart rate is sampled at approximately 9 Hz. The collection protocol
specifies 12 activities spanning static postures, locomotion, household tasks,
and exercise, including sitting, walking, cycling, vacuum cleaning, and rope
jumping. Six additional optional activities bring the full release to 18
activity labels. These recordings capture both differences among body
locations and variation in how participants perform the same activity.

\paragraph{DSADS.}
The Daily and Sports Activities dataset contains 19 activities performed by
eight participants, with five minutes recorded for each participant--activity
pair \citep{daily_and_sports_activities_256,altun2010dsads}. Five sensor units
on the torso, both arms, and both legs each provide three-axis acceleration,
angular velocity, and magnetic field measurements at 25 Hz, yielding 45
sensor channels in the original release. Activities include sitting, lying,
stair climbing, elevator travel, walking and running on a treadmill, cycling,
rowing, jumping, and basketball. Participants perform the activities in their
own style, introducing variation in movement speed and amplitude. The release
divides each five-minute recording into 60 segments of five seconds each.

\paragraph{MHealth.}
The Mobile Health dataset records ten volunteers performing 12 physical
activities using sensors attached to the chest, right wrist, and left ankle
\citep{banos2014mhealth,banos2015design}. Signals are acquired at 50 Hz and
include acceleration, angular velocity, and magnetic field measurements;
the chest device additionally records two ECG leads. The activity vocabulary
covers static postures and movements with different body involvement and
intensity, including standing, sitting, lying, walking, cycling, jogging,
running, arm elevation, and knee bending. This combination provides examples
of both localized limb movements and whole-body exercise. The original
multimodal release supplies physiological measurements alongside the motion
signals used to characterize these activities.

\paragraph{RealWorld.}
The RealWorld activity recognition dataset includes 15 participants and
eight activities: walking, running or jogging, sitting, standing, lying,
jumping, and ascending and descending stairs \citep{sztyler2016realworld}.
Devices are worn at seven positions---the chest, forearm, head, shin, thigh,
upper arm, and waist---with acceleration recorded at 50 Hz. The release also
provides gyroscope, magnetic field, GPS, light, and sound-level measurements.
Each participant performs each activity for approximately ten minutes, except
jumping, which lasts approximately 1.7 minutes because of the exertion
involved. Devices are carried using everyday attachments such as pockets and
armbands, providing variation in how activity signals are observed across
body locations.

\paragraph{UCI-HAR.}
The Human Activity Recognition Using Smartphones dataset contains recordings
from 30 volunteers aged 19--48 wearing a Samsung Galaxy S II smartphone at
the waist \citep{anguita2013ucihar,Anguita2013APD}. Its accelerometer and
gyroscope record three-axis acceleration and angular velocity at 50 Hz during
six activities: walking, ascending stairs, descending stairs, sitting,
standing, and lying. The original benchmark distributes filtered inertial
signals in 128-sample windows of 2.56 seconds with 50\% overlap, together with
engineered features, and assigns 70\% of participants to training and 30\% to
testing. The results in this paper follow the common preprocessing and
evaluation protocols described in \S\ref{sec:experiments}; the original
windowing and split describe the source benchmark.

\paragraph{USC-HAD.}
The University of Southern California Human Activity Dataset contains 14
participants performing 12 daily activities \citep{zhang2012uschad}. A
MotionNode device attached to the front of the right hip records three-axis
acceleration and angular velocity at 100 Hz, providing six motion channels.
The activities are walking forward, walking left, walking right, ascending
stairs, descending stairs, running forward, jumping, sitting, standing,
sleeping, and taking an elevator up or down. Each participant repeats each
activity five times. The activity vocabulary distinguishes movement direction
and includes both active motion and relatively stationary behaviors, providing
a benchmark for recognizing daily activities from a single body location.

\section{Implementation Details}
\label{sec:implementation}

\paragraph{Pretraining and fine-tuning settings.}
We use $P=2$ temporal bins per band and a mask ratio of $\rho=0.65$ under the
sampling protocol in \S\ref{sec:jepa}. Pretraining runs for 50 epochs with
batch size 256; fine-tuning runs for 70 epochs with batch size 64.
Both stages use AdamW with weight decay 0.05 \citep{loshchilov2019adamw}.

\paragraph{Architecture.}
The encoder uses eight Transformer blocks of width 384; the predictor uses
four blocks of width 192. Both have six attention heads, an MLP expansion
ratio of four, GELU activations, and layer normalization before each sublayer
and after the final block. Linear layers project features between these widths. Both use fixed sinusoidal position embeddings, and the predictor adds
a learned mask token.

\paragraph{Learning rates and targets.}
Pretraining and fine-tuning learning rates are updated each epoch, with linear warmup
from $10^{-6}$ over eight and ten epochs, followed by cosine decay toward
$10^{-6}$. The target encoder is initialized from the context encoder and
kept in evaluation mode with gradients disabled. EMA updates occur after each
optimizer step, with momentum increasing linearly from 0.998 toward 1.

\paragraph{Classification and channel adaptation.}
We flatten all encoder output tokens and apply layer normalization, dropout
of 0.1, and a linear classifier. The classification loss is cross-entropy
with label smoothing of 0.1. For six-channel targets,
pretrained channel embeddings are reused by index. Otherwise, the entire
channel-embedding table is reinitialized using a truncated normal distribution
($\sigma=0.02$) and learned during fine-tuning.
All other compatible encoder weights are transferred, and fixed position
embeddings are regenerated for the target token count.

\section{Additional Experiments}
\label{sec:additional_experiments}

\subsection{Regularizer ablation}
\label{sec:regularizer_ablation}

For \emph{W-JEPA (SIGReg)} and \emph{W-JEPA (VICReg)}, we keep the model and pipeline unchanged, except that we replace the original variance regularizer with SIGReg \citep{balestriero2025lejepa} or a VICReg-style variance--covariance regularizer, respectively. The JEPA prediction loss remains unchanged.

\begin{table}[htbp]
\caption{W-JEPA regularizer random-window split ablations, summarizes five seeds mean and standard deviation across the six benchmarks. Bold indicates the highest mean per metric, including ties.}
\label{tab:ablations}
\centering
\setlength{\tabcolsep}{3.5pt}
\renewcommand{\arraystretch}{1.05}
\small
\resizebox{\textwidth}{!}{%
\begin{tabular}{@{}llrrrrrrr@{}}
\toprule
Variant & Metric & PAMAP2 & DSADS & MHealth & RealWorld & UCI-HAR & USC-HAD\\
\midrule
\multirow{2}{*}{W-JEPA (SIGReg)}
  & Acc.
  & $93.62 \pm 0.07$
  & $99.02 \pm 0.09$
  & $99.23 \pm 0.27$
  & $96.92 \pm 0.06$
  & $89.74 \pm 0.30$
  & $87.76 \pm 0.45$ \\
  & F1
  & $93.70 \pm 0.04$
  & $99.01 \pm 0.10$
  & \boldsymbol{$99.22 \pm 0.24$}
  & $97.13 \pm 0.07$
  & $89.39 \pm 0.29$
  & $84.42 \pm 0.37$ \\
\midrule
\multirow{2}{*}{W-JEPA (VICReg)}
  & Acc.
  & $\mathbf{93.77 \pm 0.20}$
  & $99.11 \pm 0.08$
  & $98.97 \pm 0.57$
  & $\mathbf{97.13 \pm 0.18}$
  & $90.66 \pm 0.87$
  & $87.97 \pm 0.71$\\
  & F1
  & $\mathbf{93.82 \pm 0.22}$
  & $99.10 \pm 0.08$
  & $99.04 \pm 0.53$
  & $97.33 \pm 0.14$
  & $90.45 \pm 0.88$
  & $84.75 \pm 0.62$\\
\midrule
\multirow{2}{*}{\textbf{W-JEPA}}
& Acc. & $93.75 \pm 0.22$ & $\boldsymbol{99.22 \pm 0.13}$ & $\boldsymbol{99.28 \pm 0.35}$
& $\boldsymbol{97.13 \pm 0.14}$ & $\boldsymbol{92.91 \pm 1.23}$ & $\boldsymbol{88.54 \pm 0.35}$ \\
& F1   & $93.79 \pm 0.21$ & $\boldsymbol{99.22 \pm 0.13}$ & $99.18 \pm 0.19$
& $\boldsymbol{97.36 \pm 0.12}$ & $\boldsymbol{92.59 \pm 1.26}$ & $\boldsymbol{86.52 \pm 0.33}$ \\
  
\bottomrule
\end{tabular}%
}
\end{table}

\subsection{UCI-HAR sensitivity study}
\label{sec:uci_har_sensitivity}

\begin{figure}[ht]
    \centering
    \includegraphics[width=\textwidth]{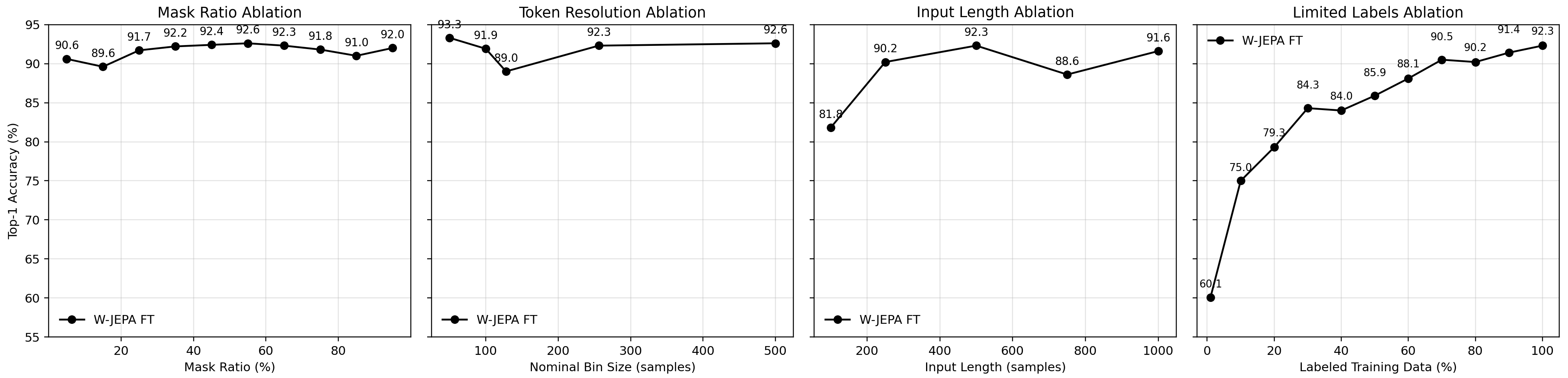}
    \caption{One-factor-at-a-time sensitivity study on UCI-HAR. From
    left to right, the panels vary the fraction of token positions masked during
    pretraining, the number of temporal bins used to summarize each wavelet
    band (shown as the nominal input samples per bin), the input-window length,
    and the requested fraction of labeled fine-tuning windows. The shared
    baseline uses a 65\% mask ratio, two bins per band, and 500-sample windows. Points report top-1 test
    accuracy from one run; nominal bin sizes are not literal wavelet receptive fields.}
    \label{fig:uci_har_ablations}
\end{figure}

\paragraph{What is varied.}
The first three panels test sensitivity to pretraining and input-design choices,
whereas the fourth tests label efficiency after the baseline pretraining run.
The mask sweep covers 5--95\% in ten-percentage-point increments. The
temporal-bin sweep uses 1, 2, 4, 5, or 10 bins for each of the five
wavelet bands, producing 30, 60, 120, 150, or 300 tokens for the six-channel
input. Its horizontal coordinate is the nominal ratio of the 500-sample window
to the number of bins. The input-length sweep uses 100, 250, 500, 750, or 1000
samples. The label sweep uses 1\% and
10--100\% of the training pool in ten-percentage-point
increments. These experiments assess
hyperparameter and data sensitivity; they do not test whether wavelets or the
JEPA objective are necessary.

\section{Additional Analyses}
\label{sec:additional_analyses}

\subsection{Exercise-bike example details}
\label{sec:dsads_detail}

\paragraph{Signal and example selection.}
Figure~\ref{fig:activity_detail} uses the left-leg gyroscope $z$-axis
(\texttt{LL\_zgyro}) from DSADS. Panels (a) and (b) show horizontal and
vertical exercise-bike cycling from WHAR subject 7, displayed as
participant 8 in panel (c). The windows begin at 264.25 s in session 119
and 37.27 s in session 127, respectively. The raw Pearson correlation
is 0.992. Signals are resampled from 25 to 100 Hz and divided into
500-sample windows that remain within session boundaries. No amplitude
normalization or post-extraction time shifting is applied.

\subsection{Attention by timescale under input noise}
\label{sec:attention_noise}

\begin{figure}[ht]
    \centering
    \includegraphics[height=0.46\linewidth]{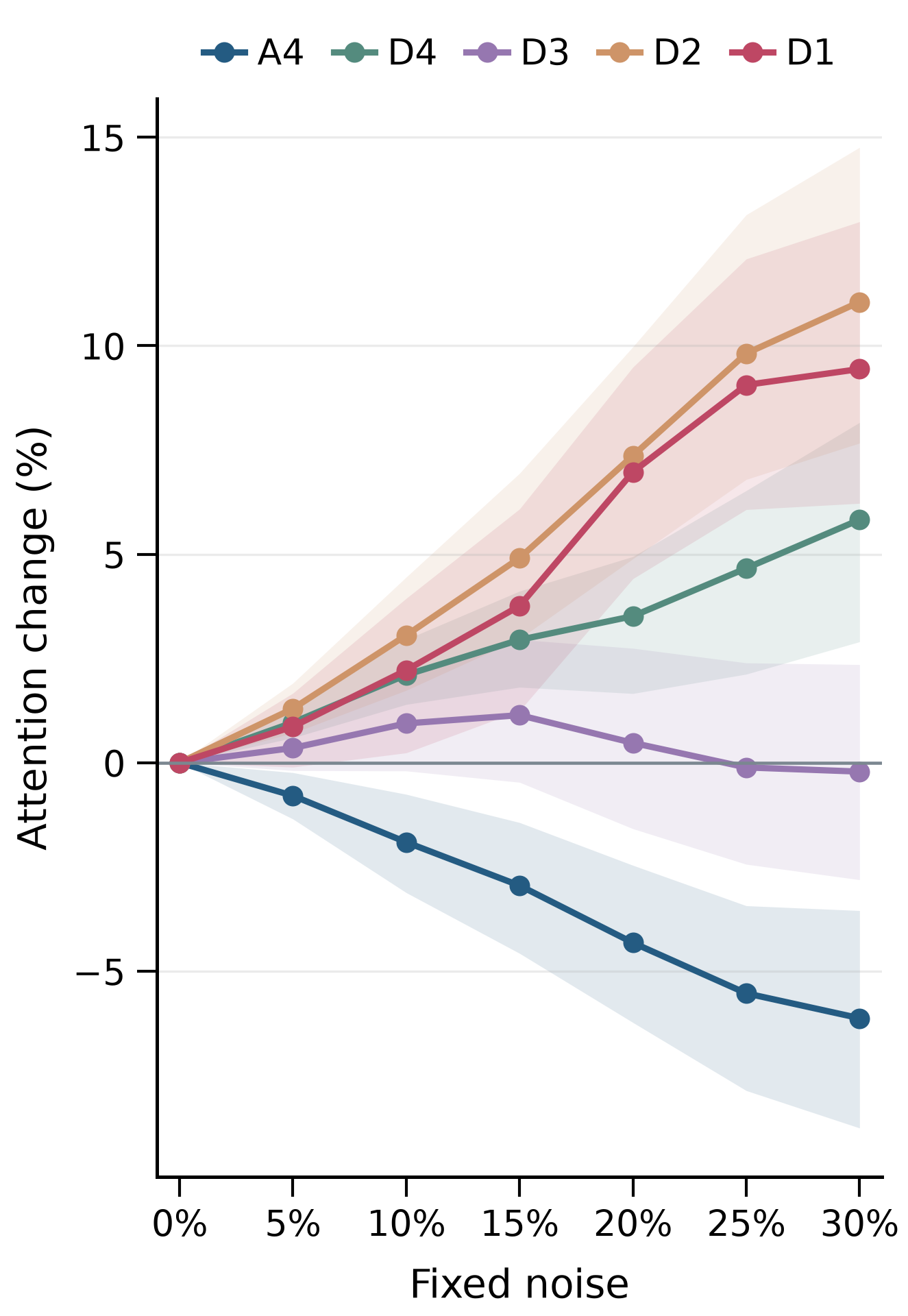}%
    \includegraphics[height=0.46\linewidth]{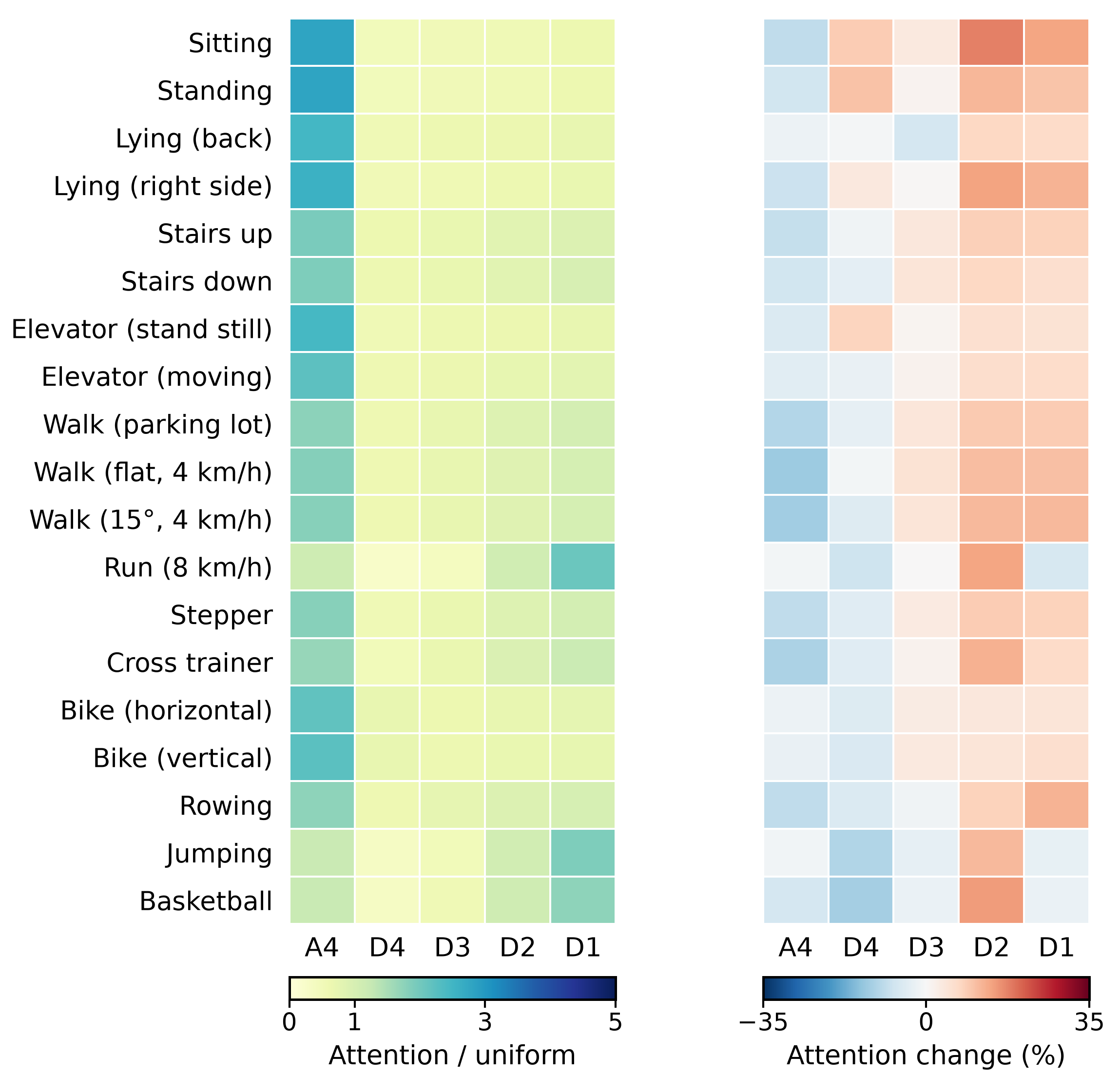}
    \caption{\textbf{Wavelet-scale attention on held-out DSADS participants.}
    Left: relative attention change across noise doses across all activities;
    Middle: clean attention by activity relative to uniform allocation.
    Right: relative attention change by activity at 30\% noise.
    All panels use the final encoder layer.}
    \label{fig:scale_attention}
\end{figure}

This section details the attention analysis summarized in
\S\ref{sec:tokenizer_analysis}. We examine how the fine-tuned encoder allocates attention across wavelet
bands on held-out DSADS participants
(Figure~\ref{fig:scale_attention}). Attention received by tokens in the
final encoder layer (layer~8) is averaged over heads and query positions,
then summed within each band. The middle heatmap divides each band's
attention by the uniform allocation of $1/5$, so a value of $1$ denotes
uniform attention.

Added noise is measured
against a fixed, training-derived per-channel dynamic RMS reference, and
attention is averaged over three different randomly generated noises per window. Both change
panels show $100(a_{\mathrm{noise}}-a_0)/a_0$, calculated after aggregating
attention within each band. Shaded area shows 95\% bootstrap intervals for the mean attention change, estimated by resampling held-out participants while keeping their clean and noisy measurements paired.

On clean inputs, $A_4$ receives the largest attention share for the activities, whereas $D_1$ receives the largest share for
running and jumping. At 30\% added noise, pooled $A_4$ attention falls
by 6.1\% relative to clean inputs, while $D_2$ and $D_1$ rise by 11.1\%
and 9.4\%.

Full-test performance remains largely intact across the tested noise
levels (Table~\ref{tab:dsads_noise_performance}). At 30\% noise, held-out
accuracy rise slightly,
while random-window accuracy and macro-F1 fall by about two points. Thus, the
attention shift toward detail bands accompanies
preserved classification performance under added noise. This is
consistent with greater use of fine-scale tokens when classifying noisy
inputs, although attention weights alone cannot establish that the shift
causes the observed robustness.

\begin{table}[t]
\centering
\caption{DSADS full-test performance under fixed input noise on a single seed. Nonzero doses report mean $\pm$ sample standard deviation across three noise realizations; clean inputs were evaluated once.}
\label{tab:dsads_noise_performance}
\begin{tabular}{@{}rcccc@{}}
\toprule
& \multicolumn{2}{c}{Held-out participants}
& \multicolumn{2}{c}{Random windows} \\
\cmidrule(lr){2-3}\cmidrule(l){4-5}
Noise (\%) & Accuracy (\%) & Macro-F1 (\%)
           & Accuracy (\%) & Macro-F1 (\%) \\
\midrule
 0 & 81.73 & 81.77 & 99.28 & 99.28 \\
 5 & $81.65 \pm 0.02$ & $81.58 \pm 0.02$
   & $99.22 \pm 0.01$ & $99.21 \pm 0.01$ \\
10 & $81.87 \pm 0.03$ & $81.67 \pm 0.02$
   & $99.05 \pm 0.04$ & $99.04 \pm 0.04$ \\
15 & $81.14 \pm 0.07$ & $80.77 \pm 0.11$
   & $98.90 \pm 0.02$ & $98.88 \pm 0.02$ \\
20 & $81.00 \pm 0.05$ & $80.73 \pm 0.08$
   & $98.66 \pm 0.04$ & $98.63 \pm 0.04$ \\
25 & $81.72 \pm 0.11$ & $81.59 \pm 0.15$
   & $98.03 \pm 0.03$ & $97.98 \pm 0.03$ \\
30 & $81.77 \pm 0.21$ & $81.60 \pm 0.26$
   & $97.14 \pm 0.09$ & $97.05 \pm 0.09$ \\
\bottomrule
\end{tabular}
\end{table}

\clearpage
\subsection{Visualization of learned representations}
\label{sec:representations}

\begin{figure}[htbp]
    \centering
    \includegraphics[width=\textwidth,height=0.82\textheight,keepaspectratio]{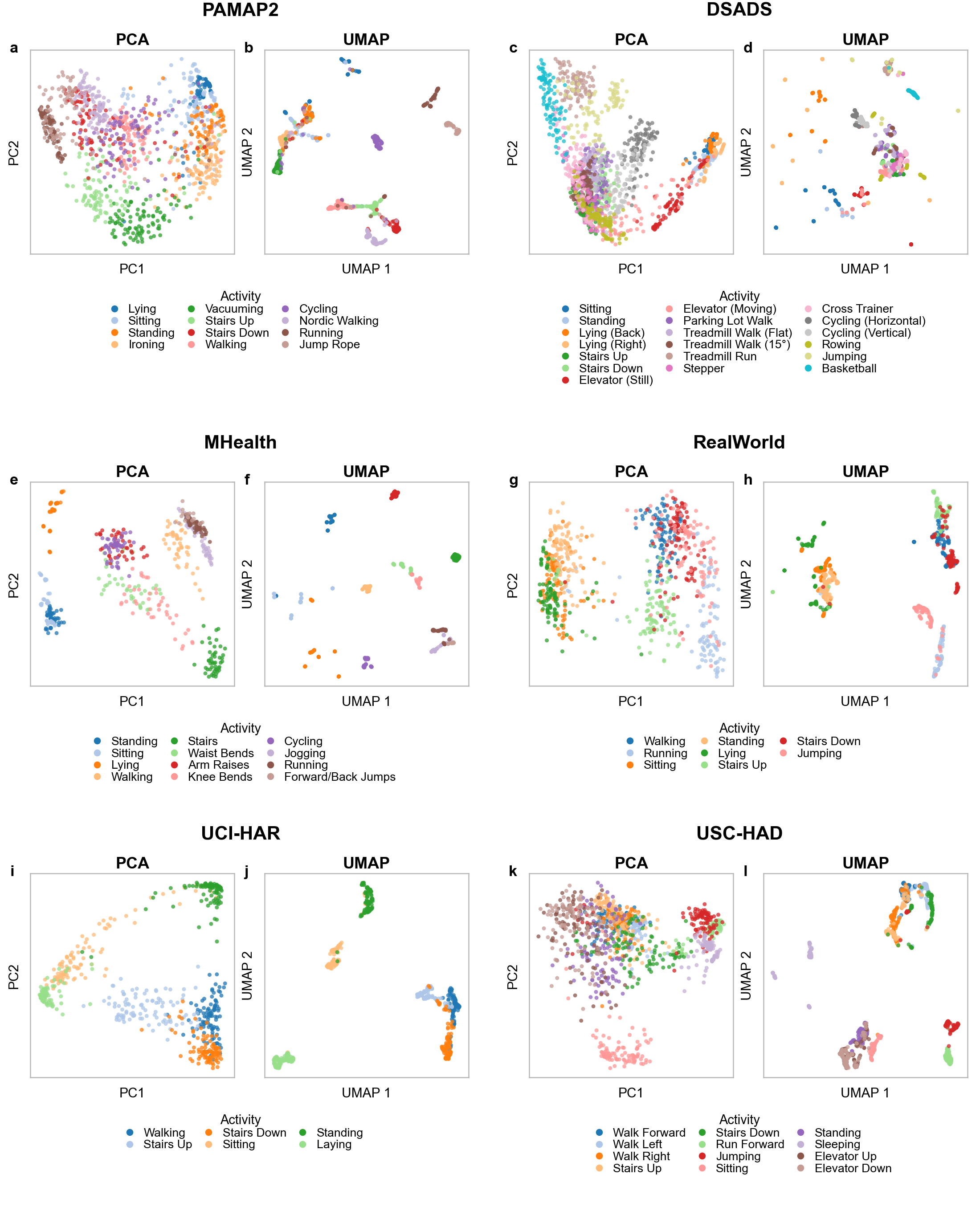}
    \caption{PCA and UMAP projections of W-JEPA’s final-block representations across six human activity recognition datasets. Each point represents a test window, colored by activity class within each dataset.}
    \label{fig:pca_umap}
\end{figure}

\end{document}